\documentclass{article}

\usepackage[preprint]{neurips_2026}

\usepackage[utf8]{inputenc} 
\usepackage[T1]{fontenc}    
\usepackage[hypertexnames=false]{hyperref} 
\usepackage{url}            
\usepackage{booktabs}       
\usepackage{amsfonts}       
\usepackage{nicefrac}       
\usepackage{microtype}      
\usepackage{xcolor}         
\usepackage[normalem]{ulem} 

\usepackage{multirow}
\usepackage{amsmath,amsfonts,bm}
\usepackage{amssymb}
\usepackage{bbm}
\usepackage{microtype}
\usepackage{booktabs} 
\usepackage{changepage,threeparttable} 
\usepackage{courier}
\usepackage{colortbl}
\usepackage{xcolor}
\usepackage{wrapfig}
\usepackage{float}
\usepackage{tcolorbox}
\definecolor{lightgrey}{rgb}{0.93, 0.93, 0.93}
\definecolor{customgreen}{HTML}{6a994e}  
\definecolor{ZEROSHOT}{HTML}{8DD2C5}
\definecolor{blue}{HTML}{03045e}
\definecolor{RAG}{HTML}{7FB2D5}
\definecolor{INTENT}{HTML}{778da9}
\definecolor{OpenSourceCodeLLM}{HTML}{D0E1F2}
\definecolor{OpenSourceGeneralLLM}{HTML}{FFC8E0}
\definecolor{ClosedSource}{HTML}{F7C97E}
\definecolor{errorcolor}{HTML}{DC4F51}
\definecolor{passcolor}{HTML}{52b69a}

\usepackage{listings}
\definecolor{codegreen}{rgb}{0,0.6,0}
\definecolor{codegray}{rgb}{0.5,0.5,0.5}
\definecolor{codepurple}{rgb}{0.58,0,0.82}
\definecolor{backcolour}{HTML}{EBEBEB}

\lstdefinestyle{mystyle}{
    backgroundcolor=\color{backcolour},   
    commentstyle=\color{codegreen},
    keywordstyle=\color{magenta},
    numberstyle=\tiny\color{codegray},
    stringstyle=\color{codepurple},
    basicstyle=\ttfamily\footnotesize,
    breakatwhitespace=false,         
    breaklines=true,                 
    captionpos=b,                    
    keepspaces=true,                 
    numbers=none,                    
    numbersep=5pt,                  
    showspaces=false,                
    showstringspaces=false,
    showtabs=false,                  
    tabsize=1,
    aboveskip=0pt, 
    belowskip=0pt, 
}

\newtcolorbox{boxD}{
    colback=sub,
    colframe=main,
    boxrule=0pt,
    toprule=6pt 
}
\definecolor{main}{HTML}{5989cf}    
\definecolor{sub}{HTML}{cde4ff}

\definecolor{blue}{rgb}{0,0,1}  
\definecolor{orange}{HTML}{FF885B}  
\definecolor{green}{HTML}{78B7D0}

\title{ORCA: Evaluating LLMs on Data Science Code Translation}

\author{
$^{1}$\textbf{Xiaolong Li}\quad
$^{1}$\textbf{Jinyang Li}\quad
$^{2}$\textbf{Bowen Qin}\quad
$^{1}$\textbf{Ge Qu}\\
$^{1}$\textbf{Nan Huo}\quad
$^{1}$\textbf{Xiaohan Xu}\quad
$^{3}$\textbf{Shipei Lin}\quad
$^{1}$\textbf{Reynold Cheng}\thanks{Corresponding author.}\\
$^{1}$The University of Hong Kong\\
$^{2}$National University of Singapore\\
$^{3}$The Chinese University of Hong Kong, Shenzhen
}

\begin{document}

\maketitle

\begin{abstract}
Data Science Code Translation (DSCT) is the process of converting code between data science libraries while preserving functional equivalence and enabling interoperability across data science ecosystems. While Large Language Models (LLMs) have demonstrated considerable progress in Data Science Code Generation (DSCG), their performance in DSCT remains insufficiently studied. To address this gap, we introduce \textsc{ORCA}, a comprehensive benchmark with two complementary settings: \textbf{\textsc{ORCA-Main}}, which comprises 1,600 carefully curated grounding-level tasks across 3 representative domains: \textbf{\textsc{Data Querying}}, \textbf{\textsc{Data Manipulation}}, and \textbf{\textsc{Deep Learning}}; and \textbf{\textsc{ORCA-Project}}, which contains 200 translation tasks over complete data science projects across 7 data science task types. Each task is accompanied by annotated reference translations and test cases for validating functional equivalence. We further incorporate a multi-stage quality verification process that thoroughly verifies task correctness and test case robustness. Experimental results demonstrate challenges in DSCT, with even frontier LLMs showing limited performance. Specifically, \texttt{Claude-Opus-4.6} achieves a success rate of 56.92\% on \textsc{ORCA-Main} and 33.67\% on \textsc{ORCA-Project}, indicating considerable room for improvement in DSCT. We also observe a clear directional preference in DSCT, where translation is consistently easier when the source code expresses the task through more explicit, fine-grained operations. Motivated by this, we propose an intent-augmented method, in which the model first infers source-code intent and then uses it as additional context for translation, achieving average absolute success-rate gains of \textbf{4.80\%} and \textbf{5.33\%} on \textsc{ORCA-Main} and \textsc{ORCA-Project}, respectively.
\end{abstract}

\section{Introduction}

Data Science Code Generation (DSCG) automates the conversion of natural language queries into executable code \citep{Lai2022DS1000AN,Yin2022NaturalLT,huang-etal-2024-da,Huang2022ExecutionbasedEF}, helping to extract valuable knowledge and insights from data \citep{10.1145/3076253, Wang2021AutoDSTH,Laradji2023CaptureTF}. This process reduces the programming barrier for data scientists and has become an integral part of decision-making and knowledge discovery across many domains \citep{han2022data,donoho201750}. 
As data scientists and engineers navigate diverse data processing and analysis frameworks, they frequently encounter scenarios that require translating code between libraries. This need arises from several critical factors: (1) \textit{Environment Migration:} Adapting to new platforms or environments often requires translating existing code to meet updated requirements \citep{cheng2025codemenv, fujii2026timemachine, shahrokhi2024pytond, schule2023blue}. For instance, a \textit{Pandas}-based analysis may need to be rewritten in SQL when the workflow must run directly over a production database. (2) \textit{Evolving Tool Ecosystems:} The rapid evolution of data science libraries continually introduces new APIs, improved execution support, and better hardware integration \citep{wang2025codesync, cox2015measuring}. Maintaining compatibility with these changes helps keep codebases up to date while reducing costly rewrites and technical debt \citep{decan2018evolution}. (3) \textit{Legacy Modernization:} Long-running data science projects often rely on outdated or unmaintained libraries throughout the codebase, requiring coordinated migration across the entire project to modernize the stack and reduce long-term maintenance burden~\citep{wang2025repotransbench,zhang2025skeleton}.

Data Science Code Translation (DSCT) addresses these challenges by automatically converting source code across data science libraries while preserving functional equivalence. At the grounding level, this requires both syntactic correctness (e.g., rewriting function calls and parameter signatures) \citep{qi2023sut,zhu2024semi} and semantic integrity (e.g., maintaining data transformation logic, indexing behavior, and computational results) \citep{huang2023program,xie2023data}. At the project level, the difficulty compounds because translation must remain consistent across interdependent files and stages such as data loading, preprocessing, modeling, and evaluation \citep{ibrahimzada2025alphatrans, ou2025rustrepotrans, meena2025transcot}. Addressing these challenges is crucial for achieving accurate and reliable code translation, enabling seamless integration and adaptation across the evolving data science landscape.

To evaluate DSCT, we introduce \textsc{ORCA}, a cr\textbf{O}ss-lib\textbf{R}ary \textbf{C}ode tr\textbf{A}nslation benchmark for data science with two complementary settings. \textsc{ORCA-Main} contains 1,600 grounding-level tasks across 3 major domains: (1) \textbf{\textsc{Data Querying}}, which translates data retrieval queries between \textit{Pandas} $\leftrightarrow$ \textit{PostgreSQL}; (2) \textbf{\textsc{Data Manipulation}}, which translates code for preprocessing and transforming data between \textit{NumPy} $\leftrightarrow$ \textit{Pandas}; and (3) \textbf{\textsc{Deep Learning}}, which translates code for model development, training, and inference between \textit{PyTorch} $\leftrightarrow$ \textit{TensorFlow}. \textsc{ORCA-Project} contains 200 translation tasks built on complete data science projects with explicit translation requirements, covering 7 representative data science task types. For each task, we manually annotate reference translations and design executable test cases to verify functional equivalence. We further conduct a multi-stage quality verification process to ensure the correctness of reference translations and the validity of test cases.


We evaluate DSCT performance across widely used LLMs on \textsc{ORCA}, including both open-source and proprietary models. Results show that even \texttt{Claude-Opus-4.6}, the strongest model in our full-benchmark evaluation, achieves success rates of only 56.92\% on \textsc{ORCA-Main} and 33.67\% on \textsc{ORCA-Project}, highlighting a clear gap between current model capabilities and the reliability required for practical deployment. Through detailed analysis, we find that DSCT difficulty is shaped by translation direction, data complexity, and cross-stage data mismatches. Using \textsc{ORCA}, we further analyze whether making source-code intent explicit can improve DSCT. We find that intent-augmented translation improves success rate by 4.80\% and 5.33\% in absolute terms at the grounding level and project level, respectively. These findings emphasize the need for continued innovation in DSCT to develop more effective and robust code translation methodologies.


\begin{figure*}[t]
    \centering
    \includegraphics[width=1.0\textwidth]{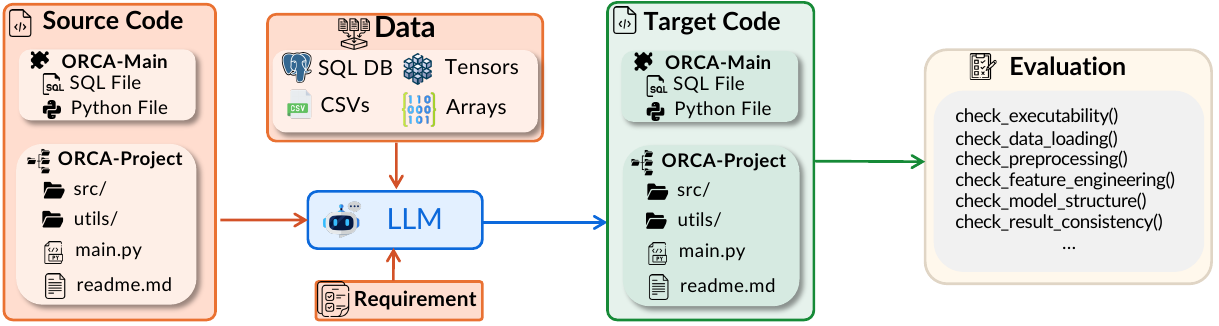}
    \caption[ORCA task format overview]{\textsc{ORCA} evaluates DSCT by providing a model with source code $\mathcal{S}$, associated data $\mathcal{D}$, and translation requirement $\mathcal{R}$, requiring translated target code $\mathcal{T}$ that is verified by execution-based test cases. \textsc{ORCA} supports both grounding-level tasks and project-level tasks.}
    \label{fig:DSCT_samples}
\end{figure*}
\section{Preliminaries}
\label{sec:task_def}

\paragraph{Task Definition.}
Data Science Code Translation (DSCT) is the process of converting code between data science libraries while preserving functional equivalence to the source code. As shown in Figure~\ref{fig:DSCT_samples}, each DSCT task is paired with associated input data $\mathcal{D}$ and a translation requirement $\mathcal{R}$. The input data may represent relational database, tabular data, tensors, or other data structures required for execution. The translation requirement specifies the source and target library settings; for project-level tasks, it also specifies the interface constraints that the translated project must satisfy. Formally, given source code $S$, associated input data $\mathcal{D}$, and translation requirement $\mathcal{R}$, a model $f_{\theta}$ produces translated code $T$:
\begin{equation}
T = f_{\theta}(S, \mathcal{D}, \mathcal{R}).
\end{equation}

We consider DSCT at two levels of granularity. In \emph{grounding-level} DSCT, each task focuses on translating a code snippet between a single source-target library pair. In \emph{project-level} DSCT, each task represents a complete data science project composed of multiple interdependent files and stages, where the source and target sides may rely on multiple data science libraries.

\paragraph{Evaluation Metric.}
To evaluate DSCT performance, we use \textbf{Success Rate} (\textbf{SR}), which corresponds to Pass@1. Unlike code generation tasks, DSCT requires exact functional preservation of a given source code, where the expected output is more deterministic. Appendix~\ref{sec:passk_cost_tradeoff} reports the cost and Pass@k performance trade-off. Let $N$ denote the total number of tasks in the evaluation set, and let $N_{\text{pass}}$ denote the number of tasks for which the translated code successfully meets all testing criteria. Formally, SR is defined as:
\begin{equation}
\mathrm{SR} = \frac{N_{\text{pass}}}{N} \times 100\%.
\end{equation}

\label{dataset_construction}
\begin{figure*}[t]
    \centering
    \includegraphics[width=1.0\textwidth]{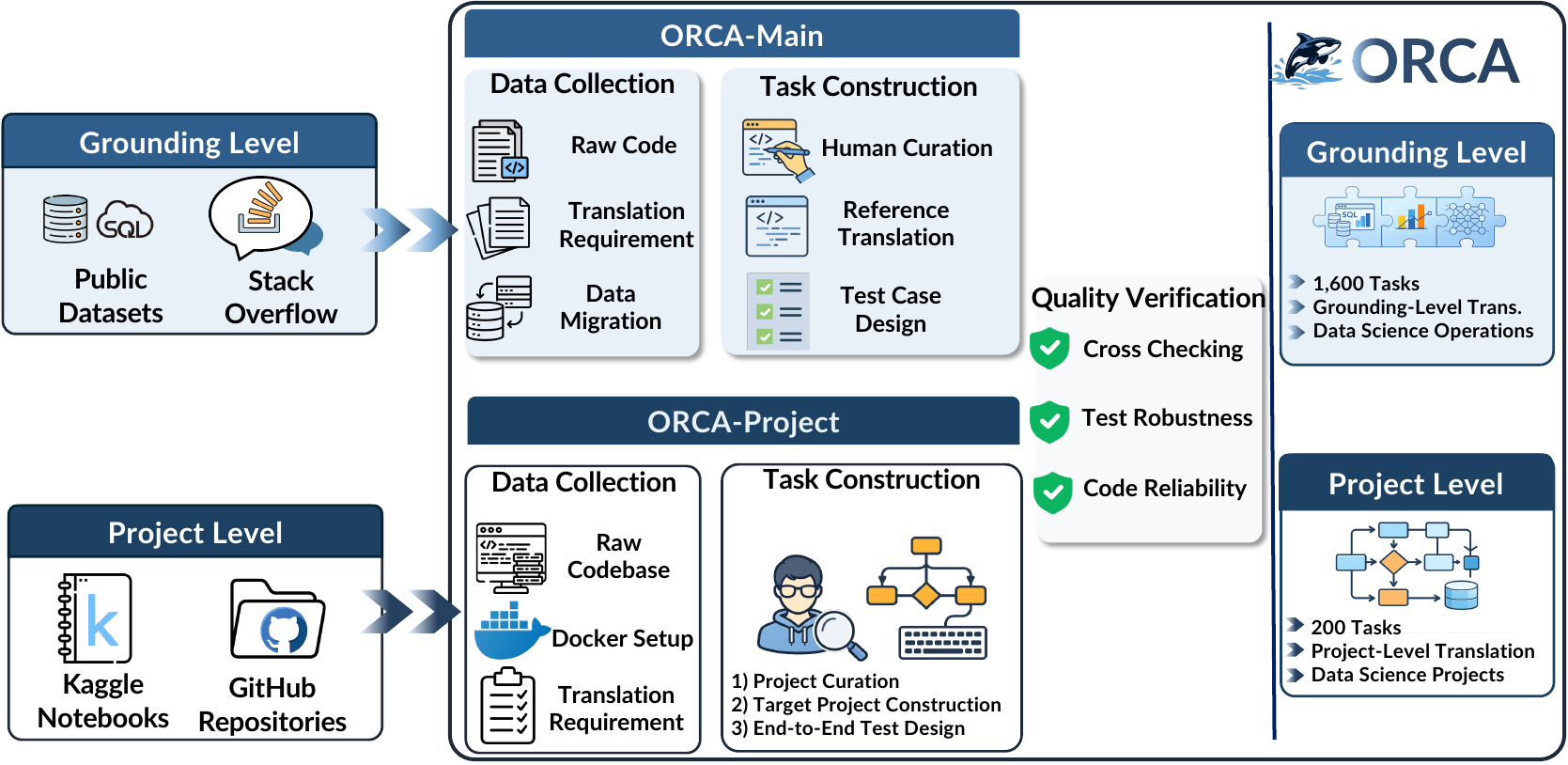}
    \caption{Overview of the ORCA construction pipeline: data collection, task construction, and quality assurance for grounding-level and project-level DSCT.}
    \label{fig:pipeline}
\end{figure*}

\section{Benchmark Construction}

This section details the construction of \textsc{ORCA}, as shown in Figure~\ref{fig:pipeline}. We first describe \textsc{ORCA-Main} (Section~\ref{sec:orca_main_construction}), then introduce \textsc{ORCA-Project} (Section~\ref{sec:orca_project_construction}), and finally present a multi-stage quality verification procedure designed to ensure the correctness and reliability of the benchmark (Section~\ref{sec:data_quality_check}).

\subsection{\textsc{ORCA-Main}}
\label{sec:orca_main_construction}

\paragraph{Source Collection and Input Preparation.}

To align \textsc{ORCA-Main} with practical data-science code translation settings, we collect grounding-level data science code from high-quality public resources. For DQ, we sample SQL queries from BIRD-SQL~\cite{Li2023CanLA} and LiveSQLBench~\cite{livesqlbench2025}, covering diverse database domains and difficulty levels. For DM and DL, we collect code snippets and associated contexts from Stack Overflow to capture realistic data manipulation and deep learning tasks raised by developers. To support execution-based evaluation, we further prepare value-equivalent inputs for both side implementations. Detailed selection criteria and data migration procedures are provided in Appendix Sections~\ref{sec:data_collection} and~\ref{sec:data_migration}, respectively.

Each task is paired with associated input data $\mathcal{D}$ and a translation requirement $\mathcal{R}$. For DQ tasks, $\mathcal{D}$ is derived from the high-quality databases provided by BIRD-SQL and LiveSQLBench. For DM and DL tasks, original Stack Overflow posts often lack complete inputs required for execution or provide only minimal illustrative values, making the code difficult to verify. To make these tasks verifiable, annotators construct the input data $\mathcal{D}$ by referring to the original thread and the task context. The translation requirement $\mathcal{R}$ specifies the source--target library direction for each task.

\paragraph{Human Curation and Reference Construction.}

After collecting source tasks, we apply 3 steps of human curation to ensure evaluation reliability, reduce contamination risk, and broaden benchmark coverage. First, to prevent inconsistent library versions in collected public code from confounding evaluation, we standardize all tasks to fixed library versions so that performance differences reflect translation difficulty rather than version-specific API changes. Second, annotators apply operation-preserving transformations to avoid directly replicating the collected source code while maintaining the same operation type being tested. For example, one aggregation operation (e.g., \texttt{MAX}) may be replaced with another (e.g., \texttt{SUM}), so that the task still evaluates aggregation operations. Finally, we enrich the benchmark coverage by modifying a subset of tasks to include data science operations that are underrepresented in the collected data, such as SQL window functions with \textcolor{blue}{\textbf{\texttt{PARTITION BY}}}, binning operations such as \textcolor{blue}{\textbf{\texttt{pd.cut}}} and \textcolor{blue}{\textbf{\texttt{np.digitize}}}, and tensor indexing operations such as \textcolor{blue}{\textbf{\texttt{torch.gather}}}.

For each curated source task $S$, annotators construct a reference translation $T$ under the corresponding $\mathcal{R}$. Instead of translating code line by line, they first identify the functionality of the source task and then reimplement it using APIs in the target library. Through this process, we retain only tasks that can be implemented in both the source and target libraries to ensure bidirectional translatability. For example, tasks relying on \texttt{take\_along\_axis} in \textit{NumPy} are removed, as \textit{Pandas} does not provide a close equivalent API for this complex indexing operation.

\paragraph{Test Case Design.}

For each translation pair, we construct executable test suites for both libraries to verify functional equivalence between the source and translated code. We design test cases and validate translation pairs through iterative refinement following a strict protocol detailed in Appendix~\ref{sec:test_case_design}.

\subsection{\textsc{ORCA-Project}}
\label{sec:orca_project_construction}

\paragraph{Source Collection and Input Preparation.}

To ground \textsc{ORCA-Project} in complete data-science workflows, we source data science projects from GitHub and Kaggle, and later curate them into 7 representative task categories. Detailed selection criteria are provided in Appendix~\ref{sec:data_collection}.

For each project-level task, the associated input data $\mathcal{D}$ is derived from the databases in BIRD-SQL~\cite{Li2023CanLA} and LiveSQLBench~\cite{livesqlbench2025}, so that project-level translations can be executed and verified under local database conditions. Prior work has shown that project-level translations are prone to interface and dependency mismatches, making them difficult to evaluate reliably with fine-grained execution-based test cases~\citep{zhang2025skeleton,wang2025repotransbench}. This challenge is further amplified for data science projects, where the task is often ambiguous and correctness depends on library-specific behavior and dependencies across multiple pipeline stages. In this case, we pair each project-level task with a skeleton-based translation requirement $\mathcal{R}$, which includes the required project structure and interface constraints that the translated project must satisfy. This design ensures project-level translations are both realistic and verifiable for reliable evaluation.

\paragraph{Human Curation and Reference Construction.}

After collecting candidate projects, annotators perform human curation to reduce contamination risk and ensure reproducibility. Annotators first adapt each project to a local database $\mathcal{D}$, selected based on schema compatibility and sufficient data scale for training and evaluation. They then reconstruct the database queries, feature definitions, and downstream code dependencies so that the codebase operates correctly under the new database context, while preserving the task type and avoiding direct reuse of the collected source implementation. For example, a time-series prediction project originally built for energy consumption data may be adapted to a local database such as \texttt{Formula\_1} to predict race lap times, while maintaining the same prediction task type. To ensure reproducibility, annotators configure a Docker environment for each task to fix library versions, system dependencies, and database services.

For each curated source project $S$, we construct a reference target project $T$ that re-implements the same data science workflow using a different set of libraries. Annotators identify the workflow stages of $S$, including data loading, preprocessing, feature engineering, modeling, and evaluation. For each stage, they list a candidate pool of target libraries. Based on the project description, annotators select libraries that fit the task's functionality, performance requirements, and domain conventions. When the selected library cannot provide a one-to-one API mapping for the required functionality, annotators first try alternative libraries from the candidate pool with closely related purposes (e.g., \textit{Pandas}, \textit{Polars}, \textit{PySpark}); if none can support the functionality, they re-implement it by composing multiple APIs from the chosen library. Projects that cannot be expressed by any library in the candidate pool are filtered out. After implementing $T$ using the selected libraries, annotators derive the translation requirement $\mathcal{R}$ from $T$ by specifying the required project structure and interface constraints. Details of the libraries used in \textsc{ORCA-Project} are provided in Appendix~\ref{sec:lib_version}, and complete annotation criteria are provided in Appendix~\ref{sec:data_collection} and \ref{sec:test_case_design}.

\paragraph{Test Case Design.}

For \textsc{ORCA-Project}, annotators construct paired test suites for each side that together verify functional equivalence at two levels. \textbf{Grounding-level tests} examine the behavior and intermediate results of individual stages such as data loading, preprocessing, and modeling, but miss interactions across stages. \textbf{Project-level tests} additionally verify cross-stage consistency and final outputs. We extend existing tests when available or construct new ones from project descriptions, and validate translation pairs through iterative refinement following a strict protocol detailed in Appendix~\ref{sec:test_case_design}.

\subsection{Multi-Stage Quality Verification}
\label{sec:data_quality_check}

After constructing \textsc{ORCA-Main} and \textsc{ORCA-Project}, we conduct a multi-stage quality verification to ensure benchmark quality. Each task is independently cross-checked by a second annotator who did not participate in its original construction. The verification covers two aspects. First, to assess \textbf{test suite robustness}, the verifier introduces intentionally incorrect implementations to confirm that the test suite catches faulty behavior, and runs valid alternative implementations to confirm that they are not falsely rejected. Second, to assess \textbf{code reliability}, the verifier designs new test cases for boundary conditions and edge cases not covered by the original suite to expose potential flaws in the annotated code.

Discrepancies between annotators are first resolved through discussion, with unresolved cases escalated to senior experts for final adjudication. Any changes made during verification, including added test cases and revised implementations, are documented with written rationales for traceability. Prior to adjudication, inter-annotator agreement is 93.2\% for grounding-level tasks and 89.7\% for project-level tasks. Tasks that remain inconsistent after adjudication are excluded from the final benchmark. The final \textsc{ORCA} consists of \textsc{ORCA-Main}, with 600 DQ tasks, 600 DM tasks, and 400 DL tasks, and \textsc{ORCA-Project}, with 200 project-level tasks.

\begin{figure}[h]
    \centering

    \includegraphics[scale=0.35]{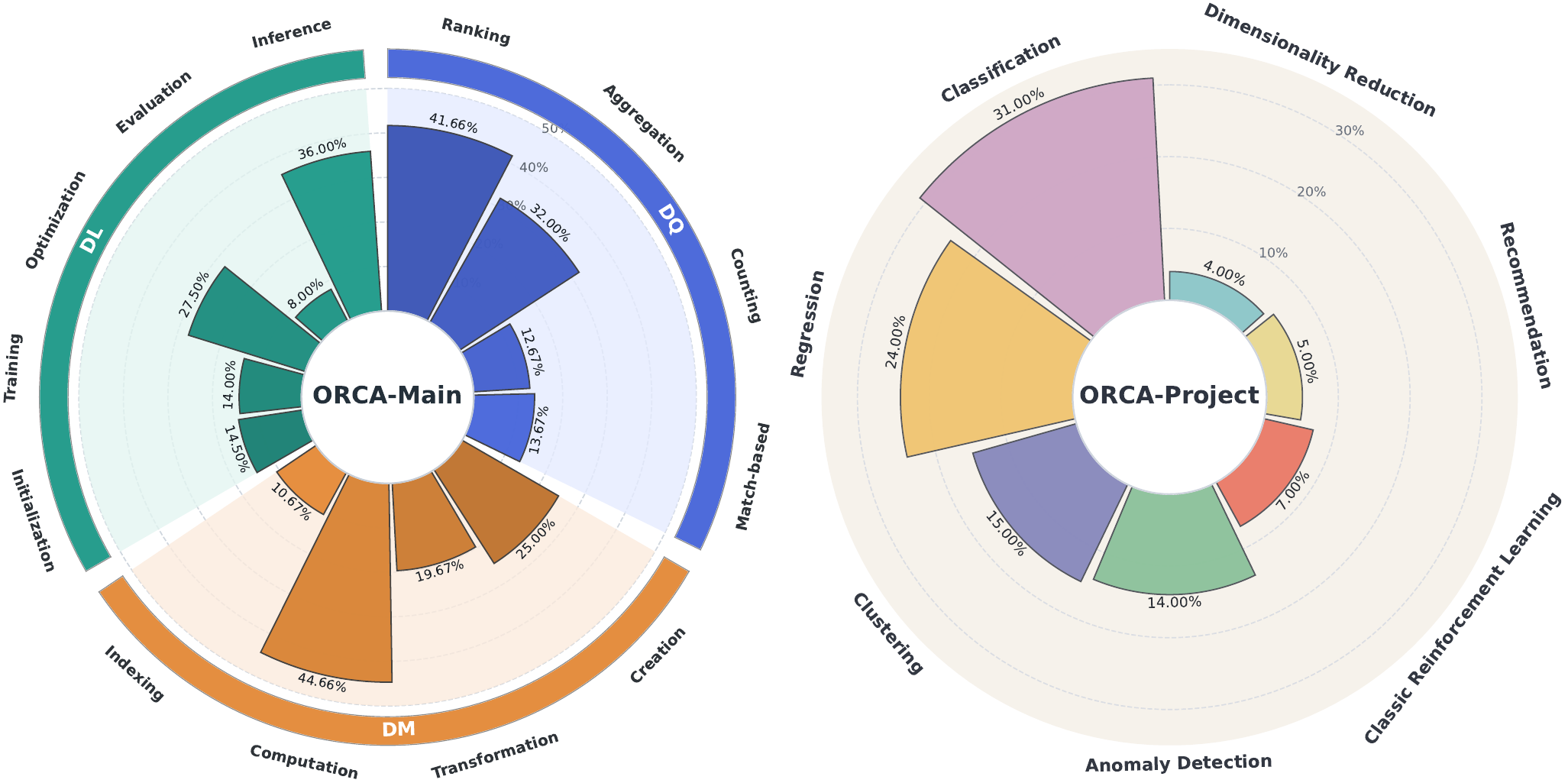}
    \caption{Distribution summary of \textsc{ORCA-Main} operation categories and \textsc{ORCA-Project} task types.}
    \label{fig:main_project_dist}
\end{figure}

\section{Dataset Statistics}
\label{sec:statistics}

\begin{wraptable}[7]{r}{0.30\textwidth}
    \vspace{-20pt}
    \centering
    \caption{Data Statistics}
    \resizebox{\linewidth}{!}{%
        {%
        \renewcommand{\arraystretch}{0.92}%
        \begin{tabular}{lcc}
            \toprule
            \textbf{\textsc{Statistic}} & \textbf{\textsc{Main}} & \textbf{\textsc{Project}}\\
            \midrule
            \textbf{Total Tasks} & 1,600 & 200 \\
            Avg. Source-Code Tokens  & 988 & 8439 \\
            \# Distinct Test Cases & 5598 & 2406 \\
            \# DS Libraries & 5 & 21 \\
            \# Distinct APIs & 496 & 615 \\
            \# Databases & 31 & 26 \\
            \midrule
            Inter-Agreement & 93.2\% & 89.7\% \\
            \bottomrule
        \end{tabular}%
        }%
    }
    \label{tab:statistic}
\end{wraptable}

Table~\ref{tab:statistic} reports the key statistics of \textsc{ORCA-Main} and \textsc{ORCA-Project}. Figure~\ref{fig:main_project_dist} summarizes the benchmark composition. For \textsc{ORCA-Main}, the left panel groups tasks by operation type across three domains. Data querying (DQ) includes ranking, aggregation, counting, and match-based queries; data manipulation (DM) covers creation, transformation, computation, and indexing; and deep learning (DL) consists of initialization, training, optimization, evaluation, and inference. The right panel shows the \textsc{ORCA-Project} distribution across project-level task categories, including classification, regression, clustering, anomaly detection, classic reinforcement learning, recommendation, and dimensionality reduction. A comparison with related benchmarks is provided in Appendix Table~\ref{tab:comparison}.

\section{Experiment}
\label{sec:experiment}

\subsection{Experimental Setup}

We evaluate a diverse set of popular open-source and strong proprietary LLMs available up to the submission on \textsc{ORCA}. For each task, models are given the source code $S$, associated input data $\mathcal{D}$, and translation requirement $\mathcal{R}$. For \textsc{ORCA-Main}, we directly prompt LLMs to generate translated code. For \textsc{ORCA-Project}, given the complexity of project-level translation, we adopt a ReAct-style agentic scaffold following previous work on general code translation~\citep{wang2025repotransbench,guan2025repotransagent}. The agent is equipped with the same tools used by our annotators, including file I/O, code generation, and program execution. We run each task three times per model and report mean Success Rate (SR) together with standard deviations across runs to capture run-to-run variation. Detailed model configurations and prompt templates are provided in Appendix~\ref{reproduce} and Appendix~\ref{sec:prompt}.

\definecolor{OpenSource}{RGB}{219, 234, 254}
\definecolor{Proprietary}{RGB}{254, 243, 199}

\begin{table}[!h]
    \vspace{-4pt}
    \caption{Performance of various LLMs on \textbf{\textsc{ORCA-Main}}. Best results in each category are marked in bold.}
  \centering
  \fontsize{10}{12}\selectfont
  \setlength{\tabcolsep}{5pt}
  \begin{tabular}{lcccc}
    \toprule
    \textbf{Model} & \textbf{Data Querying} & \textbf{Data Manipulation} & \textbf{Deep Learning} & \textbf{Overall} \\
    \midrule
    \multicolumn{5}{c}{\cellcolor{OpenSource}\textbf{Popular Open-Source Models}} \\
    Qwen2.5-Coder-7B    & $11.67_{\pm 0.93}$ & $17.22_{\pm 1.18}$  & $23.33_{\pm 1.84}$  & $16.67_{\pm 0.73}$ \\
    Llama-3.1-8B        & $7.22_{\pm 0.92}$  & $11.33_{\pm 1.09}$  & $18.50_{\pm 1.39}$  & $11.58_{\pm 0.64}$ \\
    Ministral-14B       & $18.11_{\pm 0.98}$ & $28.33_{\pm 1.36}$  & $19.33_{\pm 1.23}$  & $22.25_{\pm 0.70}$ \\
    Phi-4               & $27.78_{\pm 1.34}$ & $25.56_{\pm 0.92}$  & $25.67_{\pm 1.18}$  & $26.42_{\pm 0.67}$ \\
    Gemma-3-27B         & $20.78_{\pm 1.07}$ & $28.33_{\pm 1.30}$  & $22.00_{\pm 1.39}$  & $23.92_{\pm 0.72}$ \\
    Qwen2.5-Coder-32B   & $27.00_{\pm 0.83}$ & $33.78_{\pm 1.02}$  & $32.33_{\pm 1.81}$  & $30.88_{\pm 0.66}$ \\
    Llama-3.1-70B       & $20.89_{\pm 0.92}$ & $28.67_{\pm 1.20}$  & $30.50_{\pm 1.56}$  & $26.21_{\pm 0.69}$ \\
    MiniMax-M2-1        & $41.11_{\pm 0.98}$ & $23.56_{\pm 0.79}$  & $29.67_{\pm 1.28}$  & $31.67_{\pm 0.58}$ \\
    GLM-4.7             & $45.22_{\pm 0.79}$ & $46.89_{\pm 1.29}$  & $34.67_{\pm 1.53}$  & $43.21_{\pm 0.69}$ \\
    Qwen3-Coder-480B    & $49.00_{\pm 1.09}$ & $44.33_{\pm 0.93}$  & $39.83_{\pm 1.13}$  & $44.96_{\pm 0.61}$ \\
    DeepSeek-V3         & $52.89_{\pm 1.11}$ & $47.33_{\pm 0.88}$  & $37.50_{\pm 1.50}$  & $46.96_{\pm 0.65}$ \\
    DeepSeek-R1         & $54.56_{\pm 1.13}$ & $52.11_{\pm 1.13}$  & $32.50_{\pm 1.09}$  & $48.12_{\pm 0.66}$ \\
    Kimi-K2.5           & $57.67_{\pm 1.09}$ & $51.22_{\pm 1.08}$  & $41.17_{\pm 1.18}$  & $51.12_{\pm 0.65}$ \\
    \midrule
    \multicolumn{5}{c}{\cellcolor{Proprietary}\textbf{Strong Proprietary Models}} \\
    Claude-Sonnet-4.5   & $63.11_{\pm 0.84}$ & $50.33_{\pm 1.17}$  & $40.50_{\pm 1.39}$  & $52.67_{\pm 0.65}$ \\
    Gemini-3.1-Pro      & $68.11_{\pm 1.06}$ & $53.89_{\pm 1.08}$  & $42.67_{\pm 1.44}$  & $56.42_{\pm 0.67}$ \\
    Claude-Opus-4.6     & $\mathbf{68.67_{\pm 0.93}}$ & $\mathbf{54.00_{\pm 1.00}}$  & $\mathbf{43.67_{\pm 1.13}}$  & $\mathbf{56.92_{\pm 0.59}}$ \\
    \bottomrule
  \end{tabular}
  \vspace{-10pt}
  \label{tab:main_results}
\end{table}

\subsection{\textsc{ORCA-Main} Results}
Table~\ref{tab:main_results} reports the baseline performance of 16 LLMs on \textsc{ORCA-Main}. We observe the following: (1) \textbf{\textit{DSCT remains notably challenging}} across all evaluated models. Even the strongest model in this full-benchmark evaluation, \texttt{Claude-Opus-4.6}, achieves only 56.92\% overall, leaving substantial room for improvement. (2) \textbf{\textit{The difficulty of DSCT varies across different domains.}} DQ generally has stronger results, reflecting the relatively direct correspondence between many SQL and \textit{Pandas} operations. DM is more challenging because translating between \textit{Pandas} and \textit{NumPy} often requires reconstructing low-level array logic from high-level dataframe operations. DL also remains difficult because \textit{PyTorch} and \textit{TensorFlow} differ systematically in execution style, API conventions, and tensor behavior, so preserving functional equivalence requires more than surface-level API replacement. (3) \textbf{\textit{Improving smaller open-source LLMs for DSCT remains an important practical challenge.}} Because DSCT is inherently data-centric and often operates over sensitive datasets or private codebases, practical deployment may require local models rather than closed APIs. However, among open-source models up to 70B, the best result still reaches only 30.88\% overall.

\subsection{Directional Preferences in DSCT}
\label{sec:directional}
\begin{wrapfigure}[11]{r}{0.43\textwidth}
    \vspace{-20pt}
    \centering
    \includegraphics[width=0.96\linewidth]{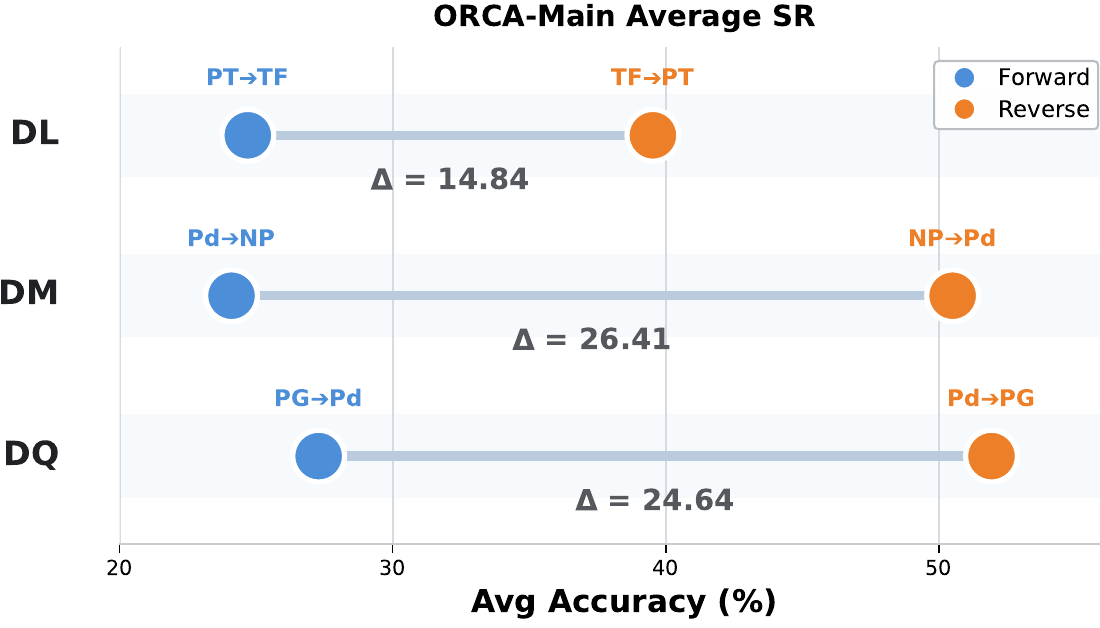}
   \caption{\small \textsc{ORCA-Main} average SR across translation directions.}
    \label{fig:directional}
\end{wrapfigure}

Figure~\ref{fig:directional} shows the average SR across all evaluated models on \textsc{ORCA-Main}, where we observe a clear directional preference in DSCT tasks. We hypothesize that this asymmetry arises when the source code $S$ expresses the task through more explicit or fine-grained operations, making it easier for models to understand the underlying functionality and translate it into the target code $T$. For each domain, one direction consistently outperforms the other direction. Specifically: \textbf{1) In DQ}, \textit{PostgreSQL} $\rightarrow$ \textit{Pandas} is consistently harder because SQL is declarative while \textit{Pandas} code is stepwise. A SQL query can express filtering, grouping, ordering, and window-based computation in a compact declarative form, whereas the corresponding \textit{Pandas} implementation often requires a sequence of intermediate operations. For example, a single SQL window clause built around \texttt{PARTITION BY} and \texttt{ORDER BY} often expands into multiple \textit{Pandas} operations, including sorting, grouping, and rolling computation. Appendix~\ref{sec:sql_dialect_appendix} reports a SQL dialect translation control experiment showing substantially smaller directional gaps when the source and target share the same declarative relational abstraction. \textbf{2) In DM}, the asymmetry is most pronounced: \textit{Pandas} $\rightarrow$ \textit{NumPy} is much harder because \textit{Pandas} provides more comprehensive data operations that need to be re-expressed through more primitive \textit{NumPy} array operations. For example, a \textit{Pandas} grouped aggregation block using \texttt{groupby().agg(\{...\})} can expand into a 19-line \textit{NumPy} implementation that manually constructs masks, slices arrays, computes reductions, and assembles the final result. \textbf{2) In DL}, with the rapid evolution of model architectures, training algorithms, and inference workflows in the AI era, both \textit{PyTorch} and \textit{TensorFlow} have developed highly integrated and comparable APIs. As a result, the directional gap is noticeably smaller than in DQ and DM. At the same time, \textit{PyTorch} $\rightarrow$ \textit{TensorFlow} remains harder than \textit{TensorFlow} $\rightarrow$ \textit{PyTorch}. One possible reason is that \textit{TensorFlow} translation often requires more framework-specific structure around computation graph construction and execution, whereas \textit{PyTorch} implementations are typically more direct and imperative. This is evidenced by the fact that \textit{TensorFlow} reference code is longer on average tokens than the corresponding \textit{PyTorch} code in our benchmark (636.08 vs.\ 497.66 tokens). A full per-model bidirectional breakdown is provided in Appendix Table~\ref{tab:bidirectional_results}.

\subsection{\textsc{ORCA-Project} Results \& Analysis}

\begin{wrapfigure}[13]{r}{0.47\textwidth}
     \vspace{-15pt}
    \centering
    \includegraphics[width=0.9\linewidth]{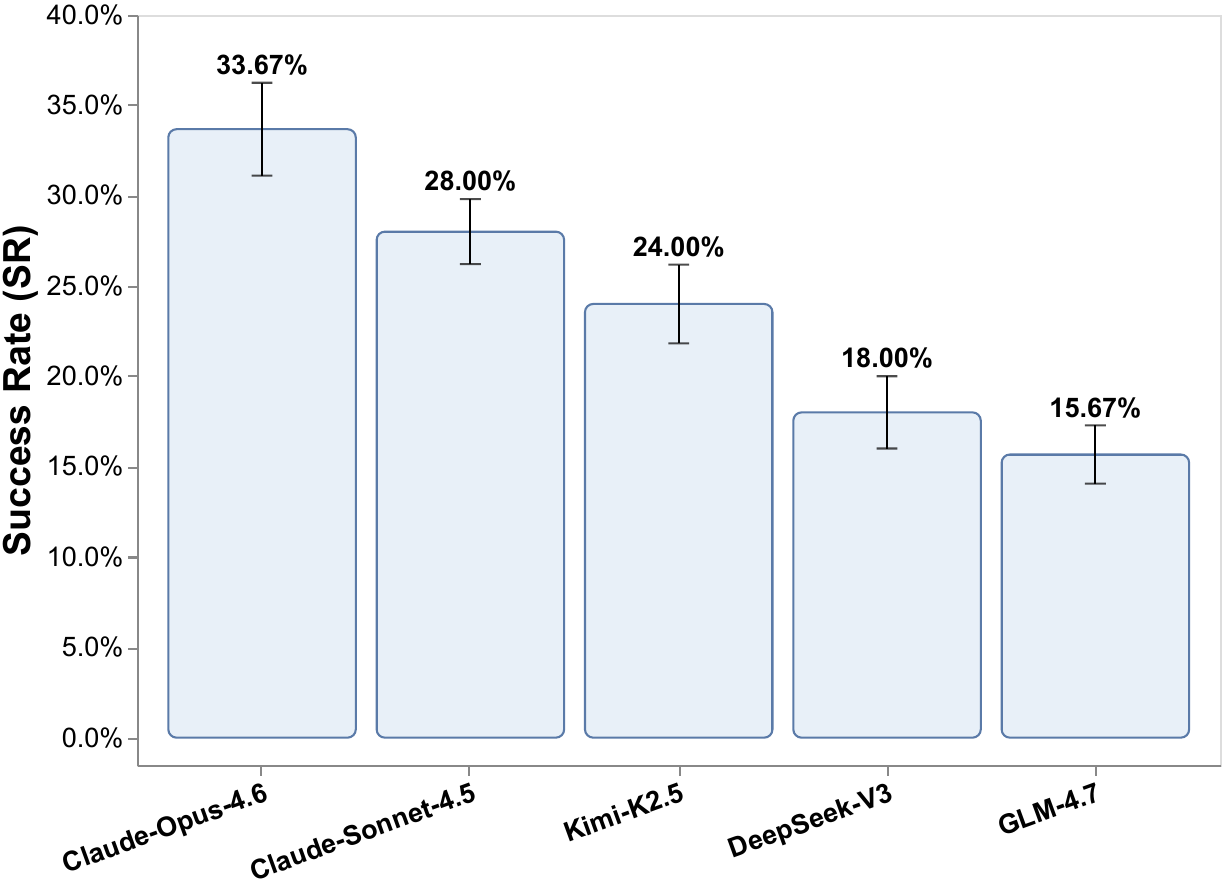}
    \caption{\small \textsc{ORCA-Project} baseline performance.}
    \label{fig:project_results}
\end{wrapfigure}

Figure~\ref{fig:project_results} shows the baseline \textsc{ORCA-Project} performance; detailed results for the models evaluated in this setting are provided in Appendix~\ref{app:project_full_results}. The strongest model in this full project-level evaluation, \texttt{Claude-Opus-4.6}, achieves only 33.67\% SR. All evaluated models show a substantial performance drop compared with \textsc{ORCA-Main}, indicating that project-level translation poses challenges beyond grounding-level operations. To better understand these difficulties, we analyze baseline results from four representative strong models, including \texttt{Claude-Opus-4.6}, \texttt{Claude-Sonnet-4.5}, \texttt{DeepSeek-V3}, and \texttt{Kimi-K2.5}, focusing on two project-level factors: data and feature complexity, and data mismatches at stage transitions.

\paragraph{Data Complexity Shapes Translation Difficulty.}

From the analyzed predictions, we find that project-level DSCT difficulty varies substantially with the underlying data complexity. Table~\ref{tab:project_complexity_bins_fields} in Appendix shows that the average SR across the four analyzed models drops from 35.23\% to 18.38\% as data complexity increases. The same pattern also holds within a single task type, with \textbf{Classification} tasks showing a decline in average SR from 36.25\% to 15.91\%, as reported in Appendix Table~\ref{tab:classification_complexity_bins_fields}. For example, a simple Classification task uses a few numeric and boolean racing features from a single table, whereas a complex Classification task joins multiple tables of trading card metadata with more than 20 fields and nested structures before training the rarity classifier. These results suggest that data complexity is an important driver of project-level DSCT difficulty.

\paragraph{Project-Level DSCT Requires Cross-Stage Data Consistency.}
Grounding-level tasks focus on single-step, localized, fine-grained data science operations, such as a query against a database, an aggregation over a tabular dataframe, or a forward pass through a neural module. In contrast, in project-level tasks, input data is processed through multiple stages in sequence: it is first loaded and cleaned, then transformed into features, and finally passed to model training or inference. Across the failed instances from the four analyzed models, 38\% involve cross-stage data mismatches, where the data produced by one stage is incompatible with what later stages expect. Typical failures include dtype mismatches, misaligned column names, incompatible tensor or array shapes, and incorrect value ordering. Mitigating such failures requires models to correctly perform a sequence of data science operations, while preserving data consistency across multiple stages.

\begin{figure*}[ht]
    \centering
    \includegraphics[width=0.97\textwidth]{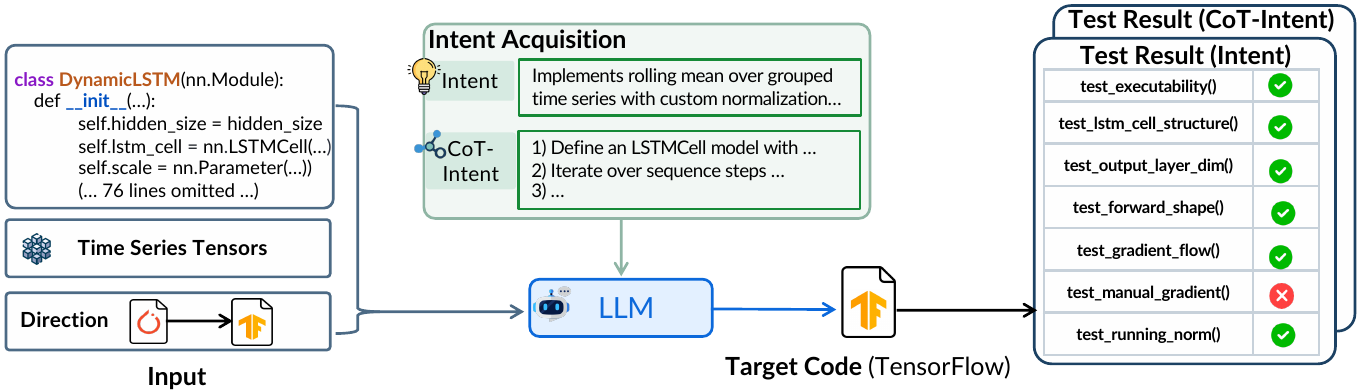}
    \caption{An example of Intent-Augmented translation from \textit{PyTorch} to \textit{TensorFlow}.}
    \label{fig:intent_rag}
\end{figure*}

\subsection{Augmenting DSCT with Code Intent}

Drawing inspiration from multilingual machine translation, where semantically equivalent sentences across languages can be mediated by a shared language-independent representation~\citep{lu2018neural,zhu2020language,vazquez2020systematic}, we investigate whether making code intent explicit can improve DSCT.

\paragraph{Implementation.}

Our Intent-Augmented method consists of two steps. (1) \textit{Intent Acquisition}: for each task, we prompt the LLM to infer the natural-language intent underlying the input code. For project-level tasks, we extend this step to build a hierarchical intent: we first generate module-level intent for each component, summarize data flow across modules, and combine them into a unified project-level intent that provides global guidance. (2) \textit{Intent-Driven Translation}: we incorporate the acquired intent as the only additional input when generating the target code, using the same model and baseline pipeline as our main experiments. A corresponding illustration is shown in Figure~\ref{fig:intent_rag}.

\begin{table*}[h]
\centering
\caption{Performance comparison across Baseline, CoT-Intent, and Intent methods on \textsc{ORCA-Main} and \textsc{ORCA-Project}. Best results in each setting are marked in bold.}
\setlength{\tabcolsep}{2.4pt}
\footnotesize
\resizebox{\linewidth}{!}{%
\begin{tabular}{@{}lcccccc@{}}
\toprule
 & \multicolumn{3}{c}{\textsc{ORCA-Main}} & \multicolumn{3}{c}{\textsc{ORCA-Project}} \\
\cmidrule(lr){2-4}\cmidrule(lr){5-7}
Model & Baseline & CoT-Intent & Intent & Baseline & CoT-Intent & Intent \\
\midrule
Qwen2.5-Coder-32B & $30.88_{\pm 0.66}$ & $29.50_{\pm 0.86}$ & $\mathbf{38.71_{\pm 0.61}}$ & $4.67_{\pm 1.04}$ & $3.33_{\pm 1.15}$ & $\mathbf{7.00_{\pm 1.32}}$ \\
Llama-3.1-70B & $26.21_{\pm 0.69}$ & $25.50_{\pm 0.72}$ & $\mathbf{33.04_{\pm 0.63}}$ & $2.00_{\pm 0.87}$ & $1.33_{\pm 0.76}$ & $\mathbf{4.00_{\pm 1.32}}$ \\
DeepSeek-V3 & $46.96_{\pm 0.65}$ & $\mathbf{48.50_{\pm 0.66}}$ & $47.38_{\pm 0.75}$ & $18.00_{\pm 2.00}$ & $23.00_{\pm 1.73}$ & $\mathbf{24.00_{\pm 2.00}}$ \\
Kimi-K2.5 & $51.12_{\pm 0.65}$ & $52.79_{\pm 0.66}$ & $\mathbf{56.04_{\pm 0.55}}$ & $24.00_{\pm 2.18}$ & $28.67_{\pm 1.89}$ & $\mathbf{31.33_{\pm 2.08}}$ \\
Claude-Opus-4.6 & $56.92_{\pm 0.59}$ & $\mathbf{62.71_{\pm 0.55}}$ & $60.92_{\pm 0.46}$ & $33.67_{\pm 2.57}$ & $\mathbf{45.00_{\pm 2.00}}$ & $42.67_{\pm 2.25}$ \\
\bottomrule
\end{tabular}
}
\label{tab:baseline_rag_intent_cot}
\end{table*}

\paragraph{Results and Analysis.}

Table~\ref{tab:baseline_rag_intent_cot} reports the performance comparison across 5 LLMs on \textsc{ORCA-Main} and \textsc{ORCA-Project}. The Intent method consistently improves over the baseline on both settings, with average absolute SR gains of 4.80\% on grounding-level tasks and 5.33\% on project-level tasks across the five evaluated models. To further validate the importance of intent for translation accuracy, we implemented a more advanced approach (\textit{CoT-Intent}) for intent generation~\citep{wei2022chain}, which prompts the model to explicitly articulate reasoning steps while generating intent. The results are model-dependent: stronger models such as \texttt{DeepSeek-V3} and \texttt{Claude-Opus-4.6} benefit from \textit{CoT-Intent}, while smaller models decline in performance. On average, \textit{CoT-Intent} produces substantially longer intent than \textit{Intent}, increasing the average intent length from 96.12 to 226.38 tokens on \textsc{ORCA-Main} and from 414.37 to 788.69 tokens on \textsc{ORCA-Project}. The additional length appears to introduce hallucinated or off-target reasoning steps that smaller models cannot reliably filter, ultimately degrading translation quality. These comparisons suggest that concise and precise intent descriptions are often more effective than longer reasoning traces.

\section{Related Work}
\paragraph{Code Translation Benchmarks.}
Early studies on code translation relied on handcrafted rules~\citep{c2rust2024,cxgo2024}, which were limited in their ability to handle complex language constructs and required substantial manual effort. Machine learning approaches utilizing aligned parallel data were later explored but were constrained by the availability and quality of such data~\citep{Koehn2007MosesOS,Nguyen2013LexicalSM,roziere2020unsupervised}. More recent work has leveraged LLMs for code translation including function-level, multilingual, and repository-level code translation~\citep{zheng2023codegeex,yang2024exploring,wang2025repotransbench,zhang2025skeleton,Bhattarai2024EnhancingCT,Rozire2021LeveragingAU,Jana2023CoTranAL}. While these benchmarks have significantly advanced the evaluation of general code translation, they primarily focus on programming-language migration and do not capture the data-centric semantics, library-specific operations, and workflow dependencies that characterize DSCT.

\paragraph{Data Science Code Benchmarks.}
Data science benchmarks have largely focused on code generation and agentic data analysis, evaluating whether models can synthesize data-centric code from natural language descriptions or task specifications~\citep{chandel2022training,Lai2022DS1000AN,huang-etal-2024-da,hu2024infiagent,jing2024dsbenchfardatascience,Chen2020HybridQAAD,tang2023ml}. These settings require models to understand diverse data structures and operations, but they do not evaluate whether models can translate existing implementations across libraries while preserving functional equivalence under framework migration. CodeTransOcean~\citep{yan2023codetransocean} includes a subset for deep learning translation, but realistic and execution-grounded evaluation across diverse data science tasks and project-level settings remains underexplored. ORCA is designed to fill this gap by covering both grounding-level and project-level DSCT under execution-based evaluation.

\section{Conclusion}

We introduce \textsc{ORCA}, a benchmark for evaluating LLMs on data science code translation at both grounding-level and project-level. \textsc{ORCA} covers realistic tasks across multiple library pairs with executable test suites for verifying functional equivalence. Experiments on widely adopted LLMs show that DSCT remains challenging. Our analyses identify three recurring sources of difficulty: directional asymmetry across library pairs, data complexity, and cross-stage data mismatches in multi-stage workflows. The proposed intent-based method improves performance at both levels, but current systems still fall short of the reliability required for practical DSCT.

\small
\bibliographystyle{plain} 
\bibliography{custom}


\appendix




\newpage
\section{Dataset Details}
\label{sec:data_details}

\subsection{Data Statistics}
\label{sec:app_data_stats}
Table~\ref{tab:comparison} further compares \textsc{ORCA} with existing data science and code translation benchmarks. Relative to prior DSCT resources, \textsc{ORCA} covers more tasks, longer and more complex code solutions, and more diverse real-world data sources. More importantly, it combines benchmark properties that are typically studied in isolation, including bidirectional translation, execution-grounded evaluation, data-science-specific cross-library translation, and both grounding-level and project-level settings. 

\begin{table}[H]
\centering
\caption{Comparison of \textsc{ORCA} with existing benchmarks. $^\dagger$Mixed refers to Stack Overflow, BIRD, and LiveSQLBench.}
\label{tab:comparison}
\resizebox{\textwidth}{!}{%
\begin{tabular}{lcccccl}
\toprule
\textbf{Benchmark} & \textbf{Level} & \textbf{\# Instances} & \textbf{\# DS Libraries} & \textbf{Avg. Test Cases / Task} & \textbf{Avg. Lines} & \textbf{Data Source} \\
\midrule
\multicolumn{7}{l}{\textit{Data Science Code Generation}} \\
DSP~\cite{chandel2022training} & Snippet & 1,119 & 8 & 2.1 & 4.5 & Notebooks \\
DS-1000~\cite{Lai2022DS1000AN} & Snippet & 1,000 & 7 & 1.6 & 3.6 & Stack Overflow \\
DA-Code~\cite{huang-etal-2024-da} & File & 500 & 10 & --- & 85 & Kaggle / GitHub \\
InfiAgent-DABench~\cite{hu2024infiagent} & File & 603 & 4 & --- & --- & GitHub (CSV) \\
DSCodeBench~\cite{jing2024dsbenchfardatascience} & Snippet & 1,000 & 10 & 200 & 22.5 & GitHub \\
\midrule
\multicolumn{7}{l}{\textit{Code Translation}} \\
TransCoder~\cite{roziere2020unsupervised} & Function & 852 & --- & 10 & 12 & GeeksForGeeks \\
HumanEval-X~\cite{zheng2023codegeex} & Function & 820 & --- & 7.7 & 8 & Hand-written \\
UniTrans~\cite{yang2024exploring} & Function & 568 & --- & 2.4 & 12.4 & GeeksForGeeks \\
CodeTransOcean (DLTrans)~\cite{yan2023codetransocean} & Snippet & 408 & 4 & --- &  37.9 & Textbook \\
RepoTransBench~\cite{wang2025repotransbench} & Repository & 1,897 & --- & 27.5 & 2394 & GitHub \\
\midrule
\multicolumn{7}{l}{\textit{Data Science Code Translation}} \\
\textbf{\textsc{ORCA-Main} (Ours)} & \textbf{Snippet} & \textbf{1,600} & \textbf{5} & \textbf{5.1} & \textbf{33.8} & \textbf{Mixed}$^\dagger$ \\
\textbf{\textsc{ORCA-Project} (Ours)} & \textbf{Repository} & \textbf{200} & \textbf{21} & \textbf{13.7} & \textbf{743.8} & \textbf{GitHub / Kaggle} \\
\bottomrule
\end{tabular}%
}
\end{table}

\subsection{Library Version}
\label{sec:lib_version}
Table~\ref{tab:lib_version} details the data science library versions used to build ORCA.
\begin{table}[H]
    \centering
    \small
    \caption{Library versions in ORCA.}
    \begin{tabular}{@{}l c l c@{}}
    \toprule
    Library & Version & Library & Version \\ \midrule
    pandas & 2.1.4 & numpy & 1.26.4 \\
    polars & 0.20.31 & pyarrow & 22.0.0 \\
    torch & 2.8.0 & tensorflow & 2.15.0 \\
    xgboost & 2.0.3 & lightgbm & 4.2.0 \\
    scikit-learn & 1.7.2 & catboost & 1.2.2 \\
    mord & 0.7 & sqlalchemy & 2.0.45 \\
    dask & 2024.1.0 & scipy & 1.13.1 \\
    statsmodels & 0.14.4 & PostgreSQL & 14.0.2 \\
    hdbscan & 0.8.33 & imitation & 1.0.0\\
    prefixspan & 0.5.2 & mlxtend & 0.23.0 \\
    pyod & 1.1.2 & \\
    \bottomrule
    \end{tabular}

    \label{tab:lib_version}
\end{table}
\subsection{Data Sources} All collected data come from resources with appropriate licenses:
\label{sec:dl_source}
\begin{itemize}
    \item \textbf{Stack Overflow:} \url{https://stackoverflow.com/} (licensed under CC BY-SA 4.0)
    \item \textbf{BIRD-SQL} \citep{Li2023CanLA}: \url{https://bird-bench.github.io/} (licensed under CC BY-SA 4.0)
    \item \textbf{LiveSQLBench} \citep{livesqlbench2025}: \url{https://livesqlbench.ai/} (licensed under CC BY-SA 4.0)
    \item \textbf{GitHub:} \url{https://github.com/} (selected repositories with explicit open-source licenses, including MIT, Apache-2.0, BSD-2-Clause, BSD-3-Clause, GPL-3.0, LGPL-3.0, and AGPL-3.0)
    \item \textbf{Kaggle public notebooks:} \url{https://www.kaggle.com/} (public notebook code licensed under Apache 2.0)
\end{itemize}

\begin{figure}[H]
    \centering
  \includegraphics[width=1.0\textwidth]{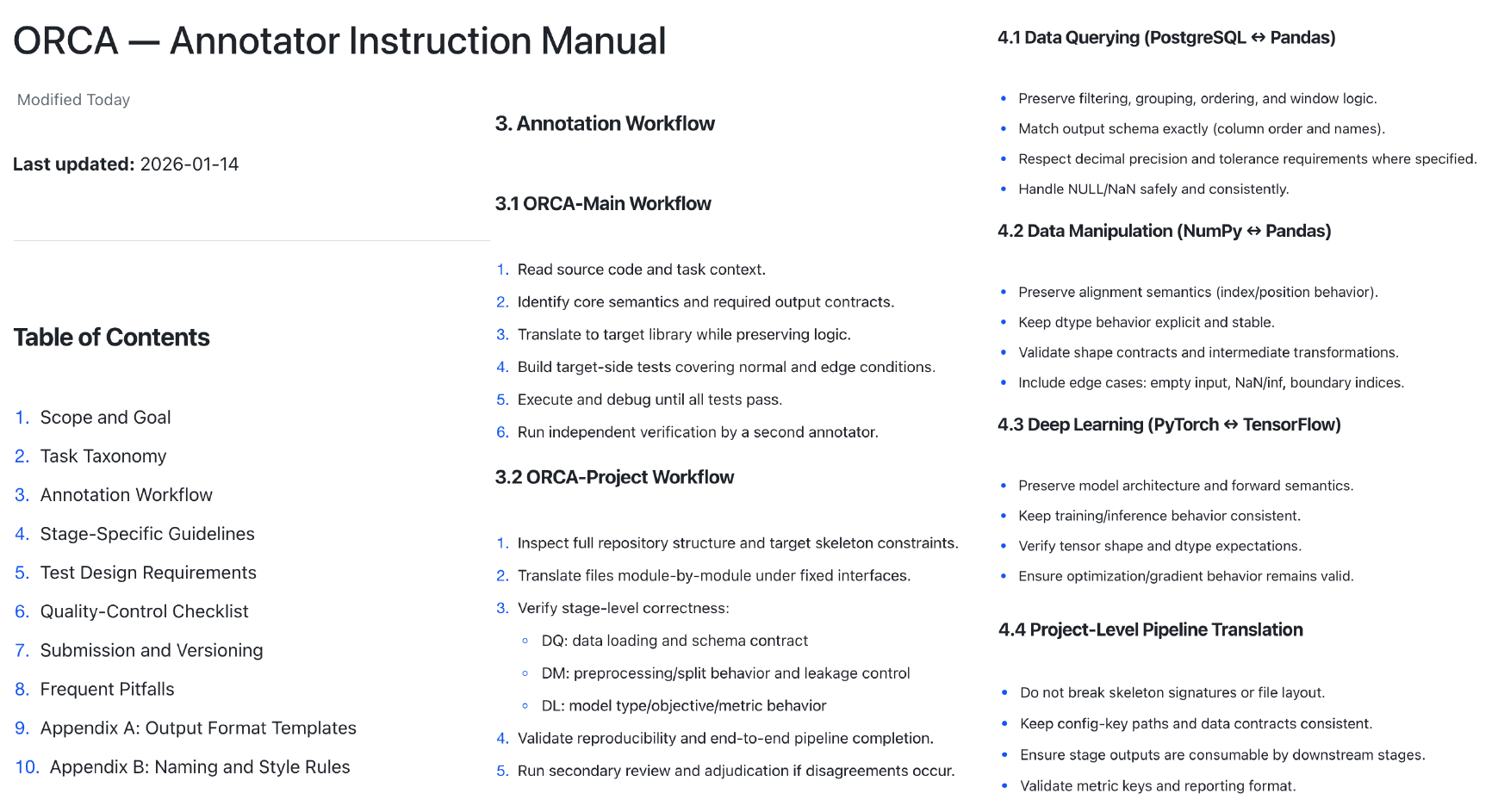}
  \caption{Screenshot of the annotator instruction manual used in the tutorial.}
  \label{fig:annotation}
\end{figure}

\section{Annotation and Evaluation Protocol}
\label{sec:anno_detail}
This section summarizes data collection, annotator qualification, test design, and compensation for ORCA construction.

\subsection{Annotator Training \& Qualification}
\label{sec:annotator_qual}

To guarantee consistent and high-quality annotations, every annotator completed a structured onboarding program \textbf{before} contributing to \textsc{ORCA}. The process comprised three stages:

\begin{enumerate}
    \item \textbf{Entrance quiz} -- a multiple-choice test covering SQL, \textit{NumPy}/\textit{Pandas} idioms, and deep learning framework semantics. Only candidates meeting the threshold advanced.
    \item \textbf{Hands-on tutorial} -- a workshop that walked annotators through
          \begin{itemize}
              \item local database / Python-environment setup,
              \item common pitfalls when mapping APIs across libraries,
              \item writing high-quality test cases to verify the translation, and
              \item the benchmark’s quality check pipeline.
          \end{itemize}
    \item \textbf{Qualification task} -- representative translation items including querying, manipulation, and deep learning domains. Annotators were required to (i) produce correct translations, (ii) provide additional hidden tests, and (iii) explain their reasoning. Submissions were spot-checked by senior reviewers before approval.
\end{enumerate}

Only qualified annotators were permitted to contribute to the benchmark. Figure~\ref{fig:annotation} shows the instruction manual used in the tutorial.

\subsection{Data Collection}
\label{sec:data_collection}

\paragraph{\textsc{ORCA-Main}.}
The collection of \textsc{ORCA-Main} starts from grounding-level code drawn from high-quality public resources, followed by manual curation. The DQ subset is built on top of two existing SQL benchmarks, BIRD-SQL~\cite{Li2023CanLA} and LiveSQLBench~\cite{livesqlbench2025}. Annotators inspect each candidate query and keep those that span a broad range of database domains and difficulty levels. The DM and DL subsets are sourced from Stack Overflow, where we filter posts by library-specific tags (\texttt{pandas} and \texttt{numpy} for DM; \texttt{pytorch} and \texttt{tensorflow} for DL). Each retrieved post must have a net score of at least 1, contain at least one accepted answer, and carry enough surrounding context to remove ambiguity in the underlying task. We further discard posts that target superficial code fixes, such as syntax errors, import issues, argument misuse, or API typos, rather than substantive data science operations.

\paragraph{\textsc{ORCA-Project}.}
The collection of \textsc{ORCA-Project} targets end-to-end data science workflows rather than isolated code snippets. We crawl candidate repositories and notebooks from GitHub and Kaggle, and apply a popularity threshold of more than 50 stars or votes to filter out low-quality or abandoned projects. Annotators then manually examine each remaining candidate, discarding exploratory notebooks, tutorial-style walkthroughs, and pipelines that terminate before reaching a trained model. The projects that survive this review implement the full workflow from data loading and preprocessing through modeling and evaluation. 

\subsection{Data Migration}
\label{sec:data_migration}
After collecting and curating the source-side tasks for \textsc{ORCA-Main}, annotators migrate the associated input data to the corresponding target-side data format, ensuring that the source and target implementations operate on value-equivalent inputs. For DQ, each PostgreSQL table is exported to a Parquet file that stores an equivalent dataframe; for DM, pandas DataFrames are converted to NumPy arrays when the source is pandas, and vice versa. For DL, annotators convert tensors between PyTorch and TensorFlow formats according to the translation direction. For \textsc{ORCA-Project}, we use the same database sources as the DQ subset to provide executable project inputs.

\subsection{Test Case Design}
\label{sec:test_case_design}

\paragraph{Grounding-Level Test Design.}
\textsc{ORCA-Main} does not rely on naive result matching. For DQ translations, each task defines explicit evaluation constraints (\texttt{order}, \texttt{distinct}, \texttt{decimal}, and optional \texttt{tolerance}) so that correctness depends on relational semantics, numeric precision, and output-order requirements. For DM and DL translations, each task includes executable framework-paired test functions (\textit{Pandas}/\textit{NumPy} and \textit{PyTorch}/\textit{TensorFlow}) that verify schema consistency, dtype correctness, aggregation or numerical behavior, and edge-case robustness. For DQ tasks, tests verify result-set cardinality, column values, and ordering. For DM tasks, tests additionally check computation logic, intermediate results, numerical precision, and data-structure consistency (e.g., shape and dtype), including challenging cases such as NaN/inf values, empty arrays, and boundary conditions. For DL tasks, tests focus on loss convergence, train/eval mode behavior, and model architecture integrity, including layer types and activation functions.

We validate each translation pair through an iterative process. If any test case fails during execution, annotators inspect whether the failure is caused by the code or the test suite and revise the corresponding component accordingly. We allow up to 5 refinement iterations and retain only those pairs for which both implementations pass all associated tests.

\paragraph{Project-Level Test Design.}
\textsc{ORCA-Project} evaluates project-level workflow translation with task-level \texttt{test\_pipeline.py} suites. The tests jointly validate data ingestion and schema integrity (DQ), preprocessing correctness including split behavior and leakage control (DM), and model-level behavior including model-type consistency and metric validity (DL), followed by reproducibility and full pipeline execution checks. Among collected projects, 46\% include test cases in the original repository or notebook, which annotators modify to fit the adapted project and extend to cover additional edge cases in the input data $\mathcal{D}$. For the remaining projects without existing tests, annotators construct new test cases based on the project description and codebase.

We validate each project-level translation using an iterative refinement procedure similar to \textsc{ORCA-Main}. When the codebase is modified during refinement, annotators update the translation requirement $\mathcal{R}$ accordingly. Since project-level translation involves multiple interdependent stages, we allow up to 10 refinement iterations and retain only those tasks for which both implementations pass all associated tests.

\subsection{Annotation Effort}
\label{sec:annotation_effort}
The annotation process required substantial expert effort to ensure the accuracy and quality of the translations. The team included 6 annotators for initial construction and 3 senior experts who oversaw the verification process. Each grounding-level translation took approximately 20 minutes to complete, with an additional 15 minutes needed for verification per task. For project-level translation, the average annotation time was around 1.5 hours per task, with an additional 1 hour for verification. The increased time for project-level tasks reflects the complexity of multi-file integration, cross-stage data consistency, and end-to-end testing. Overall, the annotation effort accumulated to approximately 1,400 person-hours across all tasks.

\subsection{Compensation}
\label{sec:annotation_compensation}
All annotators were recruited from leading universities and had between 5 to 10 years of professional experience in data science. Their academic backgrounds ensured a high level of familiarity with computational tasks and data-driven analyses, thereby contributing to the quality and reliability of the annotations. All annotators received detailed task instructions before beginning the annotation process and were compensated at a rate of USD 24 per hour for their annotation and verification work.

\section{Case Studies} \label{case_study}

This section provides representative DQ, DM, and DL case studies to illustrate how baseline methods fail and how the Intent-Augmented variants handle representative DSCT challenges.

\begin{table}[H]
\centering
\caption{Qualitative comparison of baseline, CoT-Intent, and Intent-Augmented translations on representative DQ, DM, and DL examples.}
\tiny
\renewcommand{\arraystretch}{0.98}
\setlength{\tabcolsep}{2pt}
\begin{tabular}{>{\centering\arraybackslash}p{3.05cm}|>{\centering\arraybackslash}p{3.05cm}|>{\centering\arraybackslash}p{3.05cm}|>{\centering\arraybackslash}p{3.05cm}}
\toprule
\textbf{Source Code} & \textbf{Baseline} & \textbf{CoT-Intent} & \textbf{Intent} \\
\midrule

\begin{minipage}[t]{\linewidth}\vspace{0pt}
\textbf{DQ}
\begin{lstlisting}[language=SQL,basicstyle=\ttfamily\tiny,breaklines=true,breakatwhitespace=false,columns=fullflexible,escapeinside={(*@}{@*)}]
-- (... 18 lines above ...)
WITH TransplantData AS (...),
CenterAverages AS (...)
SELECT ...
-- (... 12 lines omitted ...)
\end{lstlisting}
\textbf{Core challenge:} multi-CTE + JSON + strict output schema.
\end{minipage}
&
\begin{minipage}[t]{\linewidth}\vspace{0pt}
\begin{lstlisting}[language=Python,basicstyle=\ttfamily\tiny,breaklines=true,breakatwhitespace=false,columns=fullflexible,escapeinside={(*@}{@*)}]
# (... 20 lines above ...)
final_df = mismatch_cases.merge(...)
final_df = final_df.sort_values(...)
final_df = final_df.assign(...round(...))
# (*@\textcolor{errorcolor}{\uwave{round}}@*)
# (... 8 lines omitted ...)
\end{lstlisting}
\textcolor{errorcolor}{\textbf{Failed}: }output-format mismatch.
\end{minipage}
&
\begin{minipage}[t]{\linewidth}\vspace{0pt}
\begin{lstlisting}[language=Python,basicstyle=\ttfamily\tiny,breaklines=true,breakatwhitespace=false,columns=fullflexible,escapeinside={(*@}{@*)}]
# (... 18 lines above ...)
def safe_json_extract(...): ...
center_averages = (...).groupby(...).mean()
result_df = (...).assign(...Decimal(...))
# (... 6 lines omitted ...)
\end{lstlisting}
\textcolor{passcolor}{\textbf{Passed}} \includegraphics[width=0.4cm]{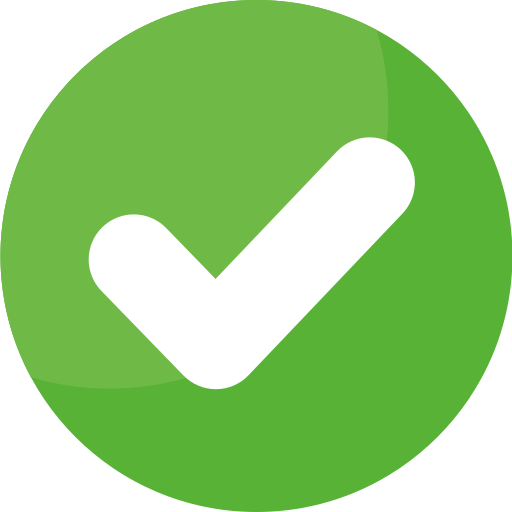}
\end{minipage}
&
\begin{minipage}[t]{\linewidth}\vspace{0pt}
\begin{lstlisting}[language=Python,basicstyle=\ttfamily\tiny,breaklines=true,breakatwhitespace=false,columns=fullflexible,escapeinside={(*@}{@*)}]
# (... 15 lines above ...)
result_df = result_df.rename(...)
result_df['avg_risk...'] = ...round(4)
# (*@\textcolor{errorcolor}{\uwave{avg\_risk}}@*)
result_df = result_df.sort_values(...)
# (... 7 lines omitted ...)
\end{lstlisting}
\textcolor{errorcolor}{\textbf{Failed}: }format/order mismatch.
\end{minipage}
\\ \hline

\begin{minipage}[t]{\linewidth}\vspace{0pt}
\textbf{DM}
\begin{lstlisting}[language=Python,basicstyle=\ttfamily\tiny,breaklines=true,breakatwhitespace=false,columns=fullflexible,escapeinside={(*@}{@*)}]
# (... 22 lines above ...)
daily_stats = groupby(...).resample('D').agg(...)
rolling_corr = groupby(...).rolling(...).corr(...)
merged_df = reset_daily_stats.merge(...)
# (... 10 lines omitted ...)
\end{lstlisting}
\textbf{Core challenge:} grouped time-index alignment.
\end{minipage}
&
\begin{minipage}[t]{\linewidth}\vspace{0pt}
\begin{lstlisting}[language=Python,basicstyle=\ttfamily\tiny,breaklines=true,breakatwhitespace=false,columns=fullflexible,escapeinside={(*@}{@*)}]
# (... 14 lines above ...)
rolling_corr = np.full((..., 31), np.nan)
...
rolling_corr[group, ts//24] = corr
# (*@\textcolor{errorcolor}{\uwave{ts//24}}@*)
# (... 6 lines omitted ...)
\end{lstlisting}
\textcolor{errorcolor}{\textbf{Failed}: }fixed day-axis causes index overflow.
\end{minipage}
&
\begin{minipage}[t]{\linewidth}\vspace{0pt}
\begin{lstlisting}[language=Python,basicstyle=\ttfamily\tiny,breaklines=true,breakatwhitespace=false,columns=fullflexible,escapeinside={(*@}{@*)}]
# (... 14 lines above ...)
daily_stats = np.zeros(..., dtype=np.float64)
result = np.zeros(..., dtype=np.float64)
result = np.round(result, 4)
# (... 6 lines omitted ...)
\end{lstlisting}
\textcolor{passcolor}{\textbf{Passed}} \includegraphics[width=0.4cm]{img/yes.png}
\end{minipage}
&
\begin{minipage}[t]{\linewidth}\vspace{0pt}
\begin{lstlisting}[language=Python,basicstyle=\ttfamily\tiny,breaklines=true,breakatwhitespace=false,columns=fullflexible,escapeinside={(*@}{@*)}]
# (... 11 lines above ...)
result = np.zeros((GROUPS, 6))
rolling_corr = np.corrcoef(...)
result = np.round(result, 4)
# (... 7 lines omitted ...)
\end{lstlisting}
\textcolor{passcolor}{\textbf{Passed}} \includegraphics[width=0.4cm]{img/yes.png}
\end{minipage}
\\ \hline

\begin{minipage}[t]{\linewidth}\vspace{0pt}
\textbf{DL}
\begin{lstlisting}[language=Python,basicstyle=\ttfamily\tiny,breaklines=true,breakatwhitespace=false,columns=fullflexible,escapeinside={(*@}{@*)}]
# (... 16 lines above ...)
class ResidualBlock(nn.Module): ...
self.shortcut = nn.Sequential(...)
class BatchSizeAware(*@\allowbreak@*)ResNet(...): ...
# (... 10 lines omitted ...)
\end{lstlisting}
\textbf{Core challenge:} shortcut + training-mode behavior.
\end{minipage}
&
\begin{minipage}[t]{\linewidth}\vspace{0pt}
\begin{lstlisting}[language=Python,basicstyle=\ttfamily\tiny,breaklines=true,breakatwhitespace=false,columns=fullflexible,escapeinside={(*@}{@*)}]
# (... 9 lines above ...)
self.shortcut = layers.Sequential(...)
self.shortcut = layers.Sequential([...])
# (*@\textcolor{errorcolor}{\uwave{layers.Sequential}}@*)
# (... 5 lines omitted ...)
\end{lstlisting}
\textcolor{errorcolor}{\textbf{Failed}: }invalid API path.
\end{minipage}
&
\begin{minipage}[t]{\linewidth}\vspace{0pt}
\begin{lstlisting}[language=Python,basicstyle=\ttfamily\tiny,breaklines=true,breakatwhitespace=false,columns=fullflexible,escapeinside={(*@}{@*)}]
# (... 20 lines above ...)
def call(self, x):
  out = self.bn2(out)
# (*@\textcolor{errorcolor}{\uwave{bn2(out)}}@*)
  out += self.shortcut(identity)
# (... 15 lines omitted ...)
\end{lstlisting}
\textcolor{errorcolor}{\textbf{Failed}: }training-state unresolved.
\end{minipage}
&
\begin{minipage}[t]{\linewidth}\vspace{0pt}
\begin{lstlisting}[language=Python,basicstyle=\ttfamily\tiny,breaklines=true,breakatwhitespace=false,columns=fullflexible,escapeinside={(*@}{@*)}]
# (... 12 lines above ...)
class ResidualBlock(tf.keras.Model):
  self.shortcut = tf.keras.Sequential([...])
def call(self, x, training=None):
  out = self.bn1(out, training=training)
# (... 8 lines omitted ...)
\end{lstlisting}
\textcolor{passcolor}{\textbf{Passed}} \includegraphics[width=0.4cm]{img/yes.png}
\end{minipage}
\\
\bottomrule
\end{tabular}
\label{tab:case_study}
\end{table}

\paragraph{DQ case.}
The DQ example requires translating a nested SQL pipeline with CTE decomposition, JSON-field extraction, and center-level aggregation into executable \textit{Pandas}. Baseline fails with output-format mismatches. CoT-Intent succeeds because its step-by-step reasoning makes the schema usage, intermediate query logic, and final output contract more explicit, helping the model preserve both the execution semantics and the required result format. In contrast, the shorter Intent variant still misses the final ordering and formatting constraints, leading to a format/order mismatch.

\paragraph{DM case.}
The DM example combines grouped daily resampling, rolling correlation, and multi-stage aggregation in a single \textit{Pandas}-to-\textit{NumPy} migration. Baseline fails on indexing logic, including fixed-axis overflow. The Intent-Augmented variant preserves aligned numeric computation and passes the full evaluation.

\paragraph{DL case.}
The DL example stresses residual-network migration details, especially shortcut behavior and training-state propagation. Baseline fails on architecture/API handling, while CoT-Intent still fails on training-state propagation and shortcut execution semantics. The Intent-Augmented translation restores the expected module structure and training-aware execution, leading to a passing result.

Across all three cases, intent-augmented prompting helps surface hidden semantic constraints that baseline prompting misses: CoT-Intent succeeds on the DQ case where strict output ordering matters, while Intent succeeds on the DM and DL cases where structural and training-state semantics dominate.

\section{Pass@k Cost Trade-off}
\label{sec:passk_cost_tradeoff}

Table~\ref{tab:passk_cost_tradeoff} compares the performance and inference cost of Pass@1 and Pass@k on \textsc{ORCA-Main} for Gemma-3-27B and DeepSeek-R1. Here, Pass@k (k=3, temperature${=}1.0$) counts a task as successful if at least one of three independently sampled outputs passes execution-based evaluation. This shows that additional sampling yields limited marginal gain ($<3\%$) on \textsc{ORCA-Main} while almost tripling the inference cost.

\begin{table}[h]
\centering
\small
\setlength{\tabcolsep}{6pt}
\renewcommand{\arraystretch}{1.10}
\caption{Performance and inference cost on \textsc{ORCA-Main}}
\label{tab:passk_cost_tradeoff}
\begin{tabular}{@{}lcccc@{}}
\toprule
\multirow{2}{*}{Model} & \multicolumn{2}{c}{Pass@1} & \multicolumn{2}{c}{Pass@k} \\
\cmidrule(lr){2-3} \cmidrule(lr){4-5}
 & SR (\%) & Cost (USD) & SR (\%) & Cost (USD) \\
\midrule
Gemma-3-27B & 23.94 & \$0.45 & 25.25 & \$1.47 \\
DeepSeek-R1 & 48.12 & \$2.66 & 50.62 & \$8.01 \\
\bottomrule
\end{tabular}
\end{table}

\section{Dialect-to-Dialect SQL Translation}
\label{sec:sql_dialect_appendix}

To contextualize the directional asymmetry observed in \textsc{ORCA-Main}, we evaluate a control setting on BIRD Mini Dev: bidirectional SQL translation among \textit{MySQL}, \textit{PostgreSQL}, and \textit{SQLite}. Table~\ref{tab:sql_dialect_bird} reports Success Rate (\%) for representative open-source models, together with a per-model average gap $\overline{|\Delta|}$ computed over the three bidirectional pairs.

We compare this control setting against \textit{Pandas}$\leftrightarrow$\textit{PostgreSQL} translation in \textsc{ORCA-Main} DQ on the same subset of seven open-source models (\texttt{Qwen2.5-Coder-7B/32B}, \texttt{Llama-3.1-8B/70B}, \texttt{Ministral-14B}, \texttt{Phi-4}, \texttt{Gemma-3-27B}). Two patterns emerge. First, dialect-to-dialect SQL translation is substantially easier in absolute terms. For each of the seven shared open-source models, we compute the mean SR over the six SQL dialect translation directions, and then macro-average these model-level means. This gives an average SR of 53.22\% in the dialect setting. In contrast, applying the same macro-averaging procedure to the corresponding \textit{Pandas}$\leftrightarrow$\textit{PostgreSQL} directions in \textsc{ORCA-Main} DQ gives an average SR of only 19.06\%. The resulting difference is 34.15 percentage points. Second, directional asymmetry is also smaller in the dialect setting. For each model, we compute the absolute SR gap for each bidirectional SQL pair, average these three gaps to obtain $\overline{|\Delta|}$, and then macro-average across the seven models. The resulting $\overline{|\Delta|}$ is 8.02 in the dialect setting, compared with 12.98 for \textit{Pandas}$\leftrightarrow$\textit{PostgreSQL} translation in \textsc{ORCA-Main} DQ. This corresponds to a reduction of 4.96 percentage points in the average bidirectional gap. Together, these results complement Section~\ref{sec:directional}: when source and target share the same declarative relational abstraction, translation is easier and less direction-dependent; when the abstraction gap widens, as in \textit{Pandas}$\leftrightarrow$\textit{PostgreSQL}, both overall difficulty and directional asymmetry increase.

\begin{table*}[!h]
\centering
\small
\setlength{\tabcolsep}{4.6pt}
\caption{SQL dialect-to-dialect translation performance on BIRD Mini Dev. Each cell is Success Rate (\%). $|\Delta|$ columns report the absolute gap between the two directions of each language pair; $\overline{|\Delta|}$ is the per-model average over the three pairs. The bottom row reports the macro-average across all listed models.}
\label{tab:sql_dialect_bird}
\resizebox{\textwidth}{!}{%
\begin{tabular}{lcc>{\columncolor{gray!10}}c cc>{\columncolor{gray!10}}c cc>{\columncolor{gray!10}}c >{\columncolor{gray!20}}c}
\toprule
& \multicolumn{3}{c}{\textbf{MySQL $\leftrightarrow$ PostgreSQL}}
& \multicolumn{3}{c}{\textbf{MySQL $\leftrightarrow$ SQLite}}
& \multicolumn{3}{c}{\textbf{PostgreSQL $\leftrightarrow$ SQLite}}
& \multicolumn{1}{c}{} \\
\cmidrule(lr){2-4} \cmidrule(lr){5-7} \cmidrule(lr){8-10}
\textbf{Model}
& My$\rightarrow$PG & PG$\rightarrow$My & $|\Delta|$
& My$\rightarrow$SQ & SQ$\rightarrow$My & $|\Delta|$
& PG$\rightarrow$SQ & SQ$\rightarrow$PG & $|\Delta|$
& $\overline{|\Delta|}$ \\
\midrule
Qwen2.5-Coder-7B   & 52.2 & 45.0 & 7.20 & 71.2 & 53.8 & 17.40 & 57.4 & 52.8 & 4.60 & 9.73 \\
Llama-3.1-8B       & 47.6 & 46.0 & 1.60 & 62.2 & 47.6 & 14.60 & 53.2 & 47.8 & 5.40 & 7.20 \\
Ministral-14B      & 51.4 & 46.0 & 5.40 & 60.2 & 47.2 & 13.00 & 52.2 & 49.4 & 2.80 & 7.07 \\
Phi-4              & 49.8 & 43.8 & 6.00 & 60.6 & 52.2 &  8.40 & 48.4 & 51.8 & 3.40 & 5.93 \\
Gemma-3-27B        & 52.0 & 48.6 & 3.40 & 68.0 & 53.2 & 14.80 & 57.6 & 52.8 & 4.80 & 7.67 \\
Qwen2.5-Coder-32B  & 51.2 & 47.6 & 3.60 & 73.0 & 53.0 & 20.00 & 56.6 & 54.0 & 2.60 & 8.73 \\
Llama-3.1-70B      & 50.4 & 44.0 & 6.40 & 65.6 & 50.6 & 15.00 & 57.6 & 49.6 & 8.00 & 9.80 \\
\midrule
\textbf{Macro-Avg} & \textbf{50.66} & \textbf{45.86} & \textbf{4.80} & \textbf{65.83} & \textbf{51.09} & \textbf{14.74} & \textbf{54.71} & \textbf{51.17} & \textbf{4.51} & \textbf{8.02} \\
\bottomrule
\end{tabular}%
}
\end{table*}

\section{\textsc{ORCA-Project} Baseline Results}
\label{app:project_full_results}

Given the substantially higher cost of project-level agentic evaluation, we evaluate a representative subset of models in this setting. Table~\ref{tab:project_full_results} reports their baseline performance on \textsc{ORCA-Project}.

\begin{table}[H]
\centering
\small
\setlength{\tabcolsep}{5.5pt}
\caption{Baseline performance of various LLMs on \textsc{ORCA-Project}. Results are reported as Success Rate (\%), with mean $\pm$ standard deviation over 3 runs. Best results are marked in bold.}
\label{tab:project_full_results}
\begin{tabular}{lc}
\toprule
\textbf{Model} & \textbf{Overall} \\
\midrule
\multicolumn{2}{c}{\cellcolor{OpenSource}\textbf{Popular Open-Source Models}} \\
Phi-4 & $0.00_{\pm 0.00}$ \\
Qwen2.5-Coder-32B & $4.67_{\pm 1.04}$ \\
Llama-3.1-70B & $2.00_{\pm 0.87}$ \\
GLM-4.7 & $15.67_{\pm 1.61}$ \\
DeepSeek-V3 & $18.00_{\pm 2.00}$ \\
Kimi-K2.5 & $24.00_{\pm 2.18}$ \\
\midrule
\multicolumn{2}{c}{\cellcolor{Proprietary}\textbf{Strong Proprietary Models}} \\
Claude-Sonnet-4.5 & $28.00_{\pm 1.80}$ \\
Claude-Opus-4.6 & $\mathbf{33.67_{\pm 2.57}}$ \\
\bottomrule
\end{tabular}
\end{table}

\section{\textsc{ORCA-Project} Difficulty by Data Complexity}
\label{app:difficulty_by_complexity}

We measure project-level data complexity with a data-only proxy that captures both the number of input sources and the schema width of the data.Specifically, for each \textsc{ORCA-Project} task, we compute the number of unique input sources multiplied by the number of input fields:
\begin{equation}
\mathrm{DataComplexity} = N_{\mathcal{D}} \times N_{\mathrm{fields}}.
\end{equation}
Here, $N_{\mathcal{D}}$ counts the distinct input sources used in the pipeline, such as database tables or Parquet files, and $N_{\mathrm{fields}}$ counts the flattened fields across these sources, including nested fields when applicable. We sort all 200 \textsc{ORCA-Project} tasks by this complexity score and divide them into three bins: \textit{Low} ($n{=}66$), \textit{Medium} ($n{=}66$), and \textit{High} ($n{=}68$). As shown in Table~\ref{tab:project_complexity_bins_fields}, SR decreases as data complexity increases, with the average SR dropping from 35.23\% in the Low bin to 18.38\% in the High bin across the evaluated models.

\begin{table}[h]
\centering
\small
\caption{SR by data-complexity bin across all \textsc{ORCA-Project} tasks. Values are reported as mean SR (\%). The last row reports the High--Low change, computed as the High-bin SR minus the Low-bin SR, with negative values indicating performance degradation.}
\label{tab:project_complexity_bins_fields}
\setlength{\tabcolsep}{4.5pt}
\begin{tabular}{lccccccc}
\toprule
\textbf{Bin} & \textbf{\#Tasks} & \textbf{Avg. Comp.} & \textbf{Opus-4.6} & \textbf{Sonnet-4.5} & \textbf{DS-V3} & \textbf{Kimi-K2.5} & \textbf{Avg.} \\
\midrule
Low    & 66 & 29.45  & 46.97 & 39.39 & 22.73 & 31.82 & 35.23 \\
Medium & 66 & 79.33  & 27.27 & 28.79 & 16.67 & 24.24 & 24.24 \\
High   & 68 & 237.79 & 26.47 & 16.18 & 14.71 & 16.18 & 18.38 \\
\midrule
\textit{High--Low Change} & -- & -- 
& \textcolor{red!70!black}{-20.50}
& \textcolor{red!70!black}{-23.21}
& \textcolor{red!70!black}{-8.02}
& \textcolor{red!70!black}{-15.64}
& \textcolor{red!70!black}{-16.85} \\
\bottomrule
\end{tabular}
\end{table}

We further analyze this trend within the same task type, focusing on \textbf{Classification}. As shown in Table~\ref{tab:classification_complexity_bins_fields}, classification tasks exhibit a similar degradation pattern as data complexity increases, with the average SR dropping from 36.25\% in the Low bin to 15.91\% in the High bin.

\begin{table*}[h]
\centering
\small
\caption{SR by data-complexity bin within \textbf{Classification} tasks. Values are reported as mean SR (\%). The last row reports the High--Low change, computed as the High-bin SR minus the Low-bin SR, with negative values indicating performance degradation.}
\label{tab:classification_complexity_bins_fields}
\setlength{\tabcolsep}{5pt}
\begin{tabular}{lccccccc}
\toprule
\textbf{Bin} & \textbf{\# Tasks} & \textbf{Avg. Comp.} & \textbf{Opus-4.6} & \textbf{Sonnet-4.5} & \textbf{DS-V3} & \textbf{Kimi-K2.5} & \textbf{Avg.} \\
\midrule
Low    & 20 & 35.60  & 50.00 & 35.00 & 20.00 & 40.00 & 36.25 \\
Medium & 20 & 78.70  & 35.00 & 30.00 & 18.00 & 35.00 & 29.50 \\
High   & 22 & 200.82 & 22.73 & 13.64 & 13.64 & 13.64 & 15.91 \\
\midrule
\textit{High--Low Change} & -- & -- 
& \textcolor{red!70!black}{-27.27}
& \textcolor{red!70!black}{-21.36}
& \textcolor{red!70!black}{-6.36}
& \textcolor{red!70!black}{-26.36}
& \textcolor{red!70!black}{-20.34} \\
\bottomrule
\end{tabular}
\end{table*}

\section{\textsc{ORCA-Main} Baseline Detailed Results}
\label{app:main_full_results}

Table~\ref{tab:bidirectional_results} reports the full bidirectional performance of all evaluated models on \textsc{ORCA-Main}, including 16 popular open-source and strong proprietary models across the DQ, DM, and DL subsets. Each cell reports the Success Rate (\%) for one translation direction, with the best result in each column within each section highlighted in bold. This detailed breakdown complements the aggregate results in Table~\ref{tab:main_results} and further supports the directional asymmetry discussed in Section~\ref{sec:directional}: across models, \textit{Pandas}$\rightarrow$\textit{PostgreSQL}, \textit{NumPy}$\rightarrow$\textit{Pandas}, and \textit{TensorFlow}$\rightarrow$\textit{PyTorch} consistently outperform their reverse directions.

\begin{table}[!h]
    \caption{Bidirectional performance on \textsc{ORCA-Main} across DQ, DM, and DL subsets. Each direction cell reports mean SR (\%) over three runs. $\Delta$ is computed as the first listed direction minus the second direction within each domain. Best SR results in each direction column within the proprietary-model section are marked in bold. Pd: Pandas, PG: PostgreSQL, NP: NumPy, TF: TensorFlow, PT: PyTorch.}
  \centering
  \fontsize{8.2}{10}\selectfont
  \setlength{\tabcolsep}{3pt}
  \resizebox{\linewidth}{!}{%
  \begin{tabular}{lccccccccc}
    \toprule
    & \multicolumn{3}{c}{\textbf{Data Querying}} & \multicolumn{3}{c}{\textbf{Data Manipulation}} & \multicolumn{3}{c}{\textbf{Deep Learning}} \\
    \cmidrule(lr){2-4} \cmidrule(lr){5-7} \cmidrule(lr){8-10}
    \textbf{Model} 
    & Pd$\rightarrow$PG & PG$\rightarrow$Pd & $\Delta$
    & NP$\rightarrow$Pd & Pd$\rightarrow$NP & $\Delta$
    & TF$\rightarrow$PT & PT$\rightarrow$TF & $\Delta$ \\
    \midrule
    \multicolumn{10}{c}{\cellcolor{OpenSource}\textbf{Popular Open-Source Models}} \\
    Qwen2.5-Coder-7B     & 15.54 & 7.80 & \textcolor{green!50!black}{+7.74} & 28.34 & 6.10 & \textcolor{green!50!black}{+22.24} & 27.74 & 18.92 & \textcolor{green!50!black}{+8.82} \\
    Llama-3.1-8B         & 9.30 & 5.14 & \textcolor{green!50!black}{+4.16} & 20.35 & 2.31 & \textcolor{green!50!black}{+18.04} & 22.77 & 14.23 & \textcolor{green!50!black}{+8.54} \\
    Ministral-14B        & 26.86 & 9.36 & \textcolor{green!50!black}{+17.50} & 42.05 & 14.61 & \textcolor{green!50!black}{+27.44} & 23.96 & 14.70 & \textcolor{green!50!black}{+9.26} \\
    Phi-4                & 35.57 & 19.99 & \textcolor{green!50!black}{+15.58} & 37.64 & 13.48 & \textcolor{green!50!black}{+24.16} & 31.93 & 19.41 & \textcolor{green!50!black}{+12.52} \\
    Gemma-3-27B          & 30.57 & 10.99 & \textcolor{green!50!black}{+19.58} & 38.54 & 18.12 & \textcolor{green!50!black}{+20.42} & 26.22 & 17.78 & \textcolor{green!50!black}{+8.44} \\
    Qwen2.5-Coder-32B    & 36.40 & 17.60 & \textcolor{green!50!black}{+18.80} & 45.20 & 22.36 & \textcolor{green!50!black}{+22.84} & 42.21 & 22.45 & \textcolor{green!50!black}{+19.76} \\
    Llama-3.1-70B        & 24.65 & 17.13 & \textcolor{green!50!black}{+7.52} & 40.93 & 16.41 & \textcolor{green!50!black}{+24.52} & 37.96 & 23.04 & \textcolor{green!50!black}{+14.92} \\
    MiniMax-M2-1         & 52.57 & 29.65 & \textcolor{green!50!black}{+22.92} & 35.82 & 11.30 & \textcolor{green!50!black}{+24.52} & 35.40 & 23.94 & \textcolor{green!50!black}{+11.46} \\
    GLM-4.7              & 60.41 & 30.03 & \textcolor{green!50!black}{+30.38} & 56.48 & 37.30 & \textcolor{green!50!black}{+19.18} & 41.31 & 28.03 & \textcolor{green!50!black}{+13.28} \\
    Qwen3-Coder-480B     & 60.73 & 37.27 & \textcolor{green!50!black}{+23.46} & 57.52 & 31.14 & \textcolor{green!50!black}{+26.38} & 46.24 & 33.42 & \textcolor{green!50!black}{+12.82} \\
    DeepSeek-V3          & 75.19 & 30.59 & \textcolor{green!50!black}{+44.60} & 67.34 & 27.32 & \textcolor{green!50!black}{+40.02} & 47.05 & 27.95 & \textcolor{green!50!black}{+19.10} \\
    DeepSeek-R1          & 77.44 & 31.68 & \textcolor{green!50!black}{+45.76} & 71.09 & 33.13 & \textcolor{green!50!black}{+37.96} & 44.08 & 20.92 & \textcolor{green!50!black}{+23.16} \\
    Kimi-K2.5            & 74.83 & 40.51 & \textcolor{green!50!black}{+34.32} & 65.90 & 36.54 & \textcolor{green!50!black}{+29.36} & 53.96 & 28.38 & \textcolor{green!50!black}{+25.58} \\
    \midrule
    \multicolumn{10}{c}{\cellcolor{Proprietary}\textbf{Strong Proprietary Models}} \\
    Claude-Sonnet-4.5    & 79.06 & 47.16 & \textcolor{green!50!black}{+31.90} & 54.64 & \textbf{46.02} & \textcolor{green!50!black}{+8.62} & 47.82 & 33.18 & \textcolor{green!50!black}{+14.64} \\
    Gemini-3.1-Pro       & 84.53 & \textbf{51.69} & \textcolor{green!50!black}{+32.84} & 71.24 & 36.54 & \textcolor{green!50!black}{+34.70} & \textbf{55.29} & 30.05 & \textcolor{green!50!black}{+25.24} \\
    Claude-Opus-4.6      & \textbf{87.27} & 50.07 & \textcolor{green!50!black}{+37.20} & \textbf{75.07} & 32.93 & \textcolor{green!50!black}{+42.14} & 48.59 & \textbf{38.75} & \textcolor{green!50!black}{+9.84} \\
    \bottomrule
  \end{tabular}%
  }
  \label{tab:bidirectional_results}
\end{table}

\section{Additional Experiments and Analyses}
\label{app:additional_analyses}

This section consolidates additional experiments and analyses that clarify the scope of data science code translation (DSCT), the interpretation of success rates under different inference settings, and the failure mechanisms observed in \textsc{ORCA}. Unless stated otherwise, all additional \textsc{ORCA-Main} experiments use the same representative subset of 300 tasks (50 tasks from each translation direction), and the project-level experiments use 100 translations formed from 50 projects evaluated in both directions.

\subsection{Positioning and Coverage of DSCT}
\label{app:dsct_positioning}

DSCT is not disjoint from general code translation or software migration; rather, it is a data-centric specialization in which the programming language may remain unchanged while the data abstraction and processing operations change. General code translation often preserves a common abstraction while rewriting syntax or language constructs. In DSCT, a correct translation must reconstruct the same computation through a different library abstraction while preserving values, indexing and aggregation behavior, schema and dtype semantics, model state, and---at the project level---contracts across pipeline stages. Accordingly, \textsc{ORCA-Main} isolates cross-library translation at the operation level, while \textsc{ORCA-Project} evaluates whether these data and computational semantics remain consistent across complete workflows. The SQL-dialect control experiment in Appendix~\ref{sec:sql_dialect_appendix} further separates syntax-level migration within a shared relational abstraction from translation across PostgreSQL and Pandas.

We selected the 5 \textsc{ORCA-Main} libraries through a systematic review of developer surveys, practical adoption, coverage of 3 core workflow stages, and support for bidirectional executable evaluation. PostgreSQL is the most desired and admired database in the 2025 Stack Overflow survey~\citep{stackoverflow2025survey}. The 2024 Python Developers Survey reports Pandas (80\%) and NumPy (75\%) as the 2 most-used tools for data exploration and processing, and PyTorch (66\%) and TensorFlow (49\%) as the 2 most-used dedicated deep-learning frameworks for model training and prediction~\citep{python2024survey}. \textsc{ORCA-Project} broadens this controlled coverage to 21 libraries and 7 project categories: classification, regression, clustering, dimensionality reduction, anomaly detection, recommendation, and classic reinforcement learning. These categories align with established machine-learning taxonomies~\citep{nist2019modernization,microsoft2024mlreference}. Ecosystems such as R and Spark and workflow components such as visualization and experiment tracking remain important extensions.

\subsubsection{Direct and non-direct operation correspondence}

To quantify cross-library expressiveness within the benchmark, we inspect every source-operation occurrence in the curated source implementation and its implementation in the paired target reference. An occurrence is \emph{direct} when 1 target operation or construct provides the same functionality, and \emph{non-direct} when the translation requires multiple target operations or lower-level reconstruction.

\begin{table}[H]
\centering
\small
\caption{Operation-level correspondence across all \textsc{ORCA-Main} tasks. Counts refer to source-operation occurrences rather than unique API names or tasks.}
\label{tab:operation_correspondence}
\setlength{\tabcolsep}{6pt}
\begin{tabular}{lrr}
\toprule
\textbf{Translation direction} & \textbf{Direct} & \textbf{Multi-step or indirect} \\
\midrule
PostgreSQL $\rightarrow$ Pandas & 2,521 (70.48\%) & 1,056 (29.52\%) \\
Pandas $\rightarrow$ PostgreSQL & 2,746 (99.35\%) & 18 (0.65\%) \\
NumPy $\rightarrow$ Pandas & 1,449 (65.80\%) & 753 (34.20\%) \\
Pandas $\rightarrow$ NumPy & 550 (42.05\%) & 758 (57.95\%) \\
PyTorch $\rightarrow$ TensorFlow & 2,665 (99.22\%) & 21 (0.78\%) \\
TensorFlow $\rightarrow$ PyTorch & 2,546 (96.66\%) & 88 (3.34\%) \\
\bottomrule
\end{tabular}
\end{table}

For DQ and DM, the easier direction also has the lower non-direct rate: Pandas$\rightarrow$PostgreSQL versus PostgreSQL$\rightarrow$Pandas, and NumPy$\rightarrow$Pandas versus Pandas$\rightarrow$NumPy. For DL, both directions have high direct-correspondence rates, consistent with the closely matched APIs of PyTorch and TensorFlow. The lower PyTorch$\rightarrow$TensorFlow SR is instead consistent with the additional TensorFlow-specific program and execution structure discussed in Section~\ref{sec:directional}.

\subsection{Task Granularity and Structural Complexity}
\label{app:granularity_complexity}

The 2 benchmark settings require different construction processes. \textsc{ORCA-Main} contains self-contained grounding tasks centered on data-science operations, whereas \textsc{ORCA-Project} contains complete repositories with cross-file structures and interface constraints. To make intermediate levels explicit, we use an LLM-assisted verifier and check its assignments using structural criteria. For \textsc{ORCA-Main}, it identifies the core source operations and functions, excluding input construction and setup code. For \textsc{ORCA-Project}, it identifies implementation files requiring translation or substantive modification, excluding tests, configuration, data files, and unchanged utilities.

\begin{table}[H]
\centering
\small
\caption{Structural granularity of \textsc{ORCA-Main} and \textsc{ORCA-Project}.}
\label{tab:structural_granularity}
\begin{tabular}{lrr@{\qquad}lrr}
\toprule
\multicolumn{3}{c}{\textbf{\textsc{ORCA-Main}}} & \multicolumn{3}{c}{\textbf{\textsc{ORCA-Project}}} \\
\cmidrule(lr){1-3}\cmidrule(lr){4-6}
\textbf{Core operations/functions} & \textbf{Tasks} & \textbf{Share} & \textbf{Files translated} & \textbf{Tasks} & \textbf{Share} \\
\midrule
1 & 41 & 2.6\% & 3 & 16 & 8.0\% \\
2--10 & 530 & 33.1\% & 4 & 127 & 63.5\% \\
11--22 & 519 & 32.4\% & 5 & 51 & 25.5\% \\
$\geq$23 & 510 & 31.9\% & 6 & 6 & 3.0\% \\
\bottomrule
\end{tabular}
\end{table}

We additionally group \textsc{ORCA-Main} tasks by domain-specific structural complexity rather than model success. DQ uses the number of relational operations, with query depth and source length as tie-breakers. DM uses the number and dependency depth of transformations, with source length as a tie-breaker. DL uses the number of framework operations, with shape-, layout-, dtype-, and device-sensitive operations and source length as tie-breakers.

\begin{table}[H]
\centering
\small
\caption{Task counts and aggregate SR across all 16 evaluated models by \textsc{ORCA-Main} structural-complexity group. Each cell reports \#tasks / SR (\%).}
\label{tab:main_complexity_breakdown}
\setlength{\tabcolsep}{8pt}
\begin{tabular}{lccc}
\toprule
\textbf{Domain} & \textbf{Simple} & \textbf{Moderate} & \textbf{Difficult} \\
\midrule
Data Querying & 180 / 45.21 & 234 / 40.68 & 186 / 33.13 \\
Data Manipulation & 189 / 39.35 & 233 / 37.50 & 178 / 34.59 \\
Deep Learning & 124 / 54.23 & 153 / 26.67 & 123 / 17.33 \\
\midrule
\textbf{Overall} & \textbf{493 / 45.23} & \textbf{620 / 36.03} & \textbf{487 / 29.67} \\
\bottomrule
\end{tabular}
\end{table}

SR decreases with structural complexity in every domain. The sharper DL decrease reflects the additional burden of shape, layout, dtype, device, and framework-state constraints. Appendix~\ref{app:difficulty_by_complexity} reports the complementary data-complexity analysis for \textsc{ORCA-Project}.

\subsection{Unified Interpretation of Evaluation Settings}
\label{app:unified_evaluation}

An \textsc{ORCA} score is conditional on the model and the inference configuration, including the prompt, tool interface, documentation access, agent harness, and interaction budget. These settings measure complementary capabilities rather than interchangeable estimates of 1 fixed number. The default \textsc{ORCA-Main} direct-prompting setting isolates atomic, operation-level translation from the provided source code, input data, and translation requirement. Its environment-driven design also supports execution-feedback and agentic evaluation. The default \textsc{ORCA-Project} setting uses a ReAct-style agent because cross-file coordination, environment interaction, and iterative debugging are integral to project-level migration. The following experiments quantify how prompts and harnesses change performance while retaining the same executable correctness criteria.

\subsubsection{GPT-family model coverage}

We first evaluate GPT-5.5 with \texttt{xhigh} reasoning and compare it with Claude-Opus-4.6 on the same representative subsets. GPT-5.5 matches or exceeds Claude-Opus-4.6 in all 6 \textsc{ORCA-Main} directions and performs better on \textsc{ORCA-Project}, but substantial failures remain and all 3 directional preferences are unchanged.

\begin{table}[H]
\centering
\small
\caption{GPT-5.5 and Claude-Opus-4.6 on representative \textsc{ORCA} subsets. Cells report passed tasks / total tasks (SR).}
\label{tab:gpt_subset_results}
\begin{tabular}{llcc}
\toprule
\textbf{Domain} & \textbf{Direction} & \textbf{GPT-5.5} & \textbf{Claude-Opus-4.6} \\
\midrule
DQ & PostgreSQL$\rightarrow$Pandas & 34/50 (68\%) & 34/50 (68\%) \\
DQ & Pandas$\rightarrow$PostgreSQL & 35/50 (70\%) & 22/50 (44\%) \\
DM & NumPy$\rightarrow$Pandas & 29/50 (58\%) & 27/50 (54\%) \\
DM & Pandas$\rightarrow$NumPy & 16/50 (32\%) & 11/50 (22\%) \\
DL & PyTorch$\rightarrow$TensorFlow & 16/50 (32\%) & 14/50 (28\%) \\
DL & TensorFlow$\rightarrow$PyTorch & 28/50 (56\%) & 22/50 (44\%) \\
\midrule
\multicolumn{2}{l}{\textbf{\textsc{ORCA-Main} overall}} & \textbf{158/300 (52.67\%)} & \textbf{130/300 (43.33\%)} \\
\multicolumn{2}{l}{\textbf{\textsc{ORCA-Project}}} & \textbf{31/100 (31\%)} & \textbf{17/100 (17\%)} \\
\bottomrule
\end{tabular}
\end{table}

\subsubsection{Functional-equivalence and library-version prompts}
\label{app:prompt_ablations}

We conduct 2 controlled ablations on the 300-task \textsc{ORCA-Main} subset using GPT-5.5 and Claude-Opus-4.6. The \emph{Detailed-Criteria} prompt explicitly lists the observable behavior to preserve. For DQ, these criteria cover returned values, columns, relational operations, row multiplicity, null and numeric behavior, and observable ordering. For DM, they cover values, shape, axes, ordering, grouping, labels, dtypes, numeric behavior, and missing or empty inputs. For DL, they cover public interfaces, tensor values and metadata, train/eval state, randomness, differentiability, gradients, and valid edge cases. The \emph{Library-Version} prompt instead specifies the exact source and target versions in Appendix Table~\ref{tab:lib_version}. Apart from the prompt addition, the experimental setting is unchanged.

\begin{table}[H]
\centering
\small
\caption{Prompt ablations on the 300-task \textsc{ORCA-Main} subset. Each cell reports SR (\%); parenthesized values are changes from the corresponding baseline.}
\label{tab:prompt_ablation_results}
\setlength{\tabcolsep}{5pt}
\begin{tabular}{llrrrr}
\toprule
\textbf{Model} & \textbf{Prompt} & \textbf{DQ} & \textbf{DM} & \textbf{DL} & \textbf{Overall} \\
\midrule
\multirow{3}{*}{GPT-5.5}
& Baseline & 69 & 45 & 44 & 52.67 \\
& Detailed-Criteria & 71 (+2) & 56 (+11) & 35 (-9) & 54.00 (+1.33) \\
& Library-Version & 68 (-1) & 48 (+3) & 37 (-7) & 51.00 (-1.67) \\
\midrule
\multirow{3}{*}{Claude-Opus-4.6}
& Baseline & 56 & 38 & 36 & 43.33 \\
& Detailed-Criteria & 58 (+2) & 42 (+4) & 39 (+3) & 46.33 (+3.00) \\
& Library-Version & 56 (0) & 40 (+2) & 39 (+3) & 45.00 (+1.67) \\
\bottomrule
\end{tabular}
\end{table}

Detailed criteria improve overall SR for both models, but the effect varies by model and domain. They state \emph{what} behavior must be preserved without supplying \emph{how} to realize it in the target library. Version strings identify the runtime environment but do not explain version-specific behavior, map source operations to target constructs, or show how to compose those constructs. The mixed version-prompt results therefore indicate that version uncertainty affects some tasks but does not explain most failures. The baseline remains a realistic human-understandable migration request: it measures whether a model can infer the behavior a user expects to preserve without requiring the user to enumerate task-specific technical criteria.

\subsubsection{Execution-feedback agent on \textsc{ORCA-Main}}

We run the ReAct-style agent used by \textsc{ORCA-Project} on the same 300 \textsc{ORCA-Main} tasks with Claude-Opus-4.6. The agent can read and write files, execute commands, and use environment feedback for up to 20 turns; final evaluation tests remain hidden.

\begin{table}[H]
\centering
\small
\caption{Direct prompting and a ReAct-style agent on \textsc{ORCA-Main}.}
\label{tab:react_main_results}
\begin{tabular}{llrrr}
\toprule
\textbf{Domain} & \textbf{Direction} & \textbf{Direct SR} & \textbf{Agent SR} & \textbf{$\Delta$} \\
\midrule
DQ & PostgreSQL$\rightarrow$Pandas & 68 & 70 & +2 \\
DQ & Pandas$\rightarrow$PostgreSQL & 44 & 58 & +14 \\
DM & NumPy$\rightarrow$Pandas & 54 & 56 & +2 \\
DM & Pandas$\rightarrow$NumPy & 22 & 24 & +2 \\
DL & PyTorch$\rightarrow$TensorFlow & 28 & 34 & +6 \\
DL & TensorFlow$\rightarrow$PyTorch & 44 & 52 & +8 \\
\midrule
\multicolumn{2}{l}{\textbf{Overall}} & \textbf{43.33} & \textbf{49.00} & \textbf{+5.67} \\
\bottomrule
\end{tabular}
\end{table}

Execution feedback improves every domain. The largest DQ gain occurs for Pandas$\rightarrow$PostgreSQL, where running both implementations can expose null handling, empty aggregation, numeric comparison, and rounding differences. DL benefits from inspecting shape, state, checkpoint, attribute, and return-type contracts. DM gains less because matching visible values may still leave the wrong container, index, dimensionality, ordering, or dtype.

\subsubsection{Frontier coding-agent evaluation}
\label{app:claude_code_eval}

We convert the same 300 \textsc{ORCA-Main} tasks and 100 \textsc{ORCA-Project} tasks to the Harbor format and follow a containerized agent-evaluation protocol based on Terminal-Bench~\citep{merrill2026terminalbench}. Each task provides the instruction, source code, input data, and an isolated Docker environment; the reference translation and final tests remain inaccessible. Claude Code uses Claude-Opus-4.6, may inspect and edit files, execute commands, receive feedback, and search online documentation. For \textsc{ORCA-Main}, its baseline is direct prompting; for \textsc{ORCA-Project}, its baseline is the paper's ReAct-style agent. All comparisons use the same underlying model and task subset.

\begin{table}[H]
\centering
\small
\caption{Claude Code performance and cost on representative \textsc{ORCA} subsets.}
\label{tab:claude_code_results}
\begin{tabular}{lrrrr}
\toprule
\textbf{Subset} & \textbf{Baseline SR} & \textbf{Claude Code SR} & \textbf{Baseline cost} & \textbf{Claude Code cost} \\
\midrule
Data Querying & 56.0 & 69.0 & -- & -- \\
Data Manipulation & 38.0 & 45.0 & -- & -- \\
Deep Learning & 36.0 & 44.0 & -- & -- \\
\textsc{ORCA-Main} overall & 43.33 & 52.67 & \$6.72 & \$202.88 \\
\textsc{ORCA-Project} & 17.0 & 26.0 & \$46.84 & \$164.89 \\
\midrule
\textbf{Total cost} & -- & -- & \textbf{\$53.56} & \textbf{\$367.77} \\
\bottomrule
\end{tabular}
\end{table}

Claude Code consistently improves both granularities. On \textsc{ORCA-Main}, trajectories show a recurring loop: reconstruct source behavior, translate, execute source and candidate implementations, and consult documentation when cross-library APIs appear similar but differ semantically. Across 300 tasks, it executes 1,358 Bash commands and accesses documentation in 292 trajectories. On \textsc{ORCA-Project}, all 100 trajectories inspect the source implementation, with 1,091 reads, 429 writes, and 250 edits; 95\% inspect database or Parquet schemas, and the agent executes 578 Bash commands. These tools help coordinate data loaders, preprocessors, models, and entry points and repair errors such as leakage, missing metrics, null propagation, and feature-name type mismatches.

The gains nevertheless come at approximately $30\times$ and $3.5\times$ the corresponding baseline costs for \textsc{ORCA-Main} and \textsc{ORCA-Project}. Agentic performance also reflects the harness, tool interfaces, exploration strategy, and turn budget. These results should therefore be interpreted as a stronger inference configuration enabled by the same benchmark, rather than as a replacement for the direct-prompting measurement. Improving both agent effectiveness and efficiency remains an important direction for DSCT.

\subsection{Systematic Failure Analysis}
\label{app:systematic_failures}

\subsubsection{Failures across prompt settings}

We classify the failed outputs from the Baseline, Detailed-Criteria, and Library-Version settings for GPT-5.5 and Claude-Opus-4.6 into 3 mutually exclusive high-level patterns using the generated translation and failed-test evidence.

\begin{table}[H]
\centering
\small
\caption{Failure patterns across the 3 prompt settings.}
\label{tab:prompt_failure_patterns}
\begin{tabular}{p{0.24\linewidth}p{0.10\linewidth}p{0.48\linewidth}}
\toprule
\textbf{Failure pattern} & \textbf{Share} & \textbf{Definition} \\
\midrule
Data or interface mismatch & 58.1\% & Incorrect columns, schema, container, public interface, shape, axis, layout, dtype, precision, or null behavior. \\
Logic or behavior mismatch & 22.9\% & Executable code with incorrect query or algorithmic logic, train/eval behavior, randomness, or gradient behavior. \\
Incomplete or invalid implementation & 19.0\% & Missing imports, helpers, or definitions; invalid target-library APIs; or failed execution. \\
\bottomrule
\end{tabular}
\end{table}

The dominant mechanism differs by domain. DQ failures concentrate on query logic (63.5\%) and numeric, type, or null behavior (33.8\%) because PostgreSQL$\leftrightarrow$Pandas crosses declarative relational queries and procedural DataFrame operations. DM failures concentrate on shape or layout (37.8\%) and output or schema interfaces (36.6\%) because Pandas uses labeled rows, columns, and indices whereas NumPy uses positional axes. DL failures concentrate on model interfaces (34.3\%) and incomplete implementations (31.9\%) because PyTorch$\leftrightarrow$TensorFlow translation must preserve classes, attributes, signatures, supporting definitions, framework state, and training behavior. Thus, DQ primarily fails on computation and query semantics, DM on data structure and representation, and DL on model interfaces and implementation completeness.

The Detailed-Criteria prompt rescues 39 baseline failures, mainly involving interfaces (15), query or computational logic (9), and type, precision, or null behavior (8), while 26 previously passing model-task pairs no longer pass. It tells the model what must be preserved but not how to implement it. The Library-Version prompt rescues 26 baseline failures and loses 26 baseline passes; the total number of failures and target-library API errors are unchanged. Across all 3 prompts, 264 of 600 matched model-task pairs (44.0\%) fail consistently, and 240 of these 264 pairs (90.9\%) retain the same fine-grained failure type. Among the persistent same-type failures, 58.8\% are data or interface mismatches, 21.7\% are logic or behavior mismatches, and 19.6\% are incomplete or invalid implementations. Prompt clarification and version metadata affect individual outcomes, but neither supplies the target-side procedure needed to construct the correct representation, interface, computation, and implementation.

\subsubsection{Claude Code failures and validation behavior}

For each failed Claude Code task, we inspect the final translated code and recorded trajectory. We compare \textsc{ORCA-Main} against direct prompting and \textsc{ORCA-Project} against the ReAct-style baseline.

\begin{table}[H]
\centering
\small
\caption{Claude Code and baseline failure patterns on \textsc{ORCA-Main}. Counts are over failed tasks in each setting.}
\label{tab:claude_code_main_failures}
\begin{tabular}{lrr}
\toprule
\textbf{Failure pattern} & \textbf{Claude Code} & \textbf{Baseline} \\
\midrule
Data or interface mismatch & 85/142 (59.9\%) & 99/170 (58.2\%) \\
Logic or behavior mismatch & 32/142 (22.5\%) & 44/170 (25.9\%) \\
Incomplete or invalid implementation & 25/142 (17.6\%) & 27/170 (15.9\%) \\
\bottomrule
\end{tabular}
\end{table}

Claude Code reduces the absolute number of failures in all 3 categories, but their relative distribution remains similar. The remaining failures preserve the domain-specific structure above: 26 of 31 DQ failures involve logic or behavior; 50 of 55 DM failures involve data or interface mismatches; and 31 of 56 DL failures involve data or interface mismatches while 19 are incomplete or invalid implementations.

\begin{table}[H]
\centering
\small
\caption{Claude Code and ReAct-baseline failure patterns on the 100-task \textsc{ORCA-Project} subset. Counts are over failed tasks in each setting.}
\label{tab:claude_code_project_failures}
\begin{tabular}{lrr}
\toprule
\textbf{Failure pattern} & \textbf{Claude Code} & \textbf{ReAct baseline} \\
\midrule
Cross-stage interface & 35/74 (47.3\%) & 35/83 (42.2\%) \\
Local data semantics & 5/74 (6.8\%) & 7/83 (8.4\%) \\
Output or project contract & 30/74 (40.5\%) & 39/83 (47.0\%) \\
Modeling, implementation, or execution & 4/74 (5.4\%) & 2/83 (2.4\%) \\
\bottomrule
\end{tabular}
\end{table}

Claude Code reduces output or project-contract failures from 39 to 30 and local-data failures from 7 to 5, while 35 cross-stage failures remain. It can repair individual cross-stage mismatches but does not solve this category reliably because running components independently does not verify that every producer preserves the schema, ordering, preprocessing state, and other assumptions of downstream consumers.

The trajectories explain this gap. Among 142 failed \textsc{ORCA-Main} tasks, the primary validation gap is that 91 execute only the candidate, 28 execute source and candidate without comparing their results, 22 compare them but miss the failing property, and 1 ends with a visible unresolved error. Among 74 failed \textsc{ORCA-Project} tasks, 73 execute only the candidate and 1 executes both without comparing them. Claude Code repairs 41 \textsc{ORCA-Main} and 12 \textsc{ORCA-Project} baseline failures, but among tasks failed by both systems, 93.0\% of \textsc{ORCA-Main} pairs and 83.1\% of \textsc{ORCA-Project} pairs retain the same high-level category. Execution and documentation help when the agent checks the relevant property; otherwise, verifying that a candidate runs does not establish equivalence to the source.

These findings motivate equivalence-aware agent validation: execute source and target implementations on matched inputs and compare rows and values for DQ; indices, axes, containers, and dtypes for DM; public interfaces, model state, shapes, and gradients for DL; and producer-consumer contracts across project stages.

\subsubsection{Project-level failure taxonomy}

We further analyze the failed \textsc{ORCA-Project} translations from the 4 models used in Section~5.4 and assign each failure to 1 mutually exclusive primary pattern.

\begin{table}[H]
\centering
\small
\caption{Systematic failure breakdown for the project-level analysis in Section~5.4.}
\label{tab:project_failure_taxonomy}
\begin{tabular}{p{0.24\linewidth}p{0.10\linewidth}p{0.48\linewidth}}
\toprule
\textbf{Failure pattern} & \textbf{Share} & \textbf{Definition} \\
\midrule
Cross-stage interface & 38\% & Incompatible schema, dtype, shape, ordering, preprocessing state, train--inference alignment, or I/O between an identifiable producer and consumer. \\
Local data semantics & 38\% & Incorrect query, transformation, split, leakage decision, feature computation, or numerical operation within 1 stage. \\
Output or project contract & 13\% & Violation of a required output schema, formatting, cardinality, file, function, class, signature, entry point, or return format. \\
Modeling, implementation, or execution & 11\% & Errors in model construction, training, prediction, loss or metric behavior, completeness, API usage, or execution. \\
\bottomrule
\end{tabular}
\end{table}

We break down the 2 most frequent patterns into multi-label subtypes. Cross-stage failures most often involve column/schema mismatches and train--inference misalignment, while local-data failures most often involve field extraction and missing-value handling. A failure can receive multiple subtype labels, so each distribution need not sum to 100\%.

\begin{table}[H]
\centering
\small
\caption{Subtype breakdowns for cross-stage and local-data failures.}
\label{tab:project_failure_subtypes}
\begin{tabular}{lr@{\qquad}lr}
\toprule
\multicolumn{2}{c}{\textbf{Cross-stage interface}} & \multicolumn{2}{c}{\textbf{Local data semantics}} \\
\cmidrule(lr){1-2}\cmidrule(lr){3-4}
\textbf{Subtype} & \textbf{Share} & \textbf{Subtype} & \textbf{Share} \\
\midrule
Column or schema & 72.44\% & Field/nested-value extraction & 56.39\% \\
Train--inference alignment & 50.67\% & Completeness/missing values & 37.44\% \\
Dtype & 39.11\% & Parsing/type conversion & 32.60\% \\
Feature or value ordering & 24.44\% & Query/filter/join/aggregation & 30.84\% \\
Serialization or stage I/O & 16.44\% & Split/sampling/leakage & 25.99\% \\
Shape or rank & 12.89\% & Feature/target transformation & 23.79\% \\
Normalization/preprocessing state & 2.22\% & & \\
\bottomrule
\end{tabular}
\end{table}

\subsection{Construction Automation and Analysis Artifacts}
\label{app:construction_automation}

\textsc{ORCA} was not generated through a centralized LLM pipeline. Scripts automate mechanical steps such as data collection, environment setup, and repeated test execution. Expert annotators select and adapt tasks, construct reference translations and executable tests, and determine functional equivalence. LLM assistance was neither required nor standardized in the annotation protocol. As described in Section~3.3 and Appendix~\ref{sec:anno_detail}, every task must pass its executable tests, be independently checked using intentionally incorrect and valid alternative implementations, and undergo senior-expert adjudication when disagreements remain. The complete process required approximately 1,400 person-hours.

GitHub/Kaggle and BIRD-SQL/LiveSQLBench serve different roles in \textsc{ORCA-Project}. GitHub and Kaggle provide realistic project structures, task logic, and end-to-end workflows, whereas BIRD-SQL and LiveSQLBench provide recent, realistic inputs for local execution. We do not directly reuse the original project--data pairs: annotators reconstruct each instance on a compatible local database while preserving the task type and underlying logic. This reduces direct project--data memorization and removes heterogeneous external data dependencies, while retaining realistic workflows and enabling reproducible execution.

The release contains the expert-annotated tests for every instance. A translation succeeds only when it passes all associated tests; test-level outcomes can therefore identify near-passing translations and the specific properties that remain incorrect. Together with the complexity, direction, and failure-type analyses above, these artifacts support interpretation beyond a single aggregate SR and enable future prompting and agentic methods to be evaluated under the same functional-equivalence standard.

\section{Model Aliases and Runtime Details} \label{model_details}

The following provider/model identifiers are used throughout the paper and evaluation code. 
\begin{itemize}
    \item \texttt{claude-opus-4-6}
    \item \texttt{claude-sonnet-4-5-20250929}
    \item \texttt{gemini-3.1-pro-preview}
    \item \texttt{qwen3-coder-480b-a35b}
    \item \texttt{deepseek.r1}
    \item \texttt{deepseek.v3}
    \item \texttt{glm-4.7}
    \item \texttt{kimi-k2.5}
    \item \texttt{minimax-m2.1}
    \item \texttt{Qwen2.5-Coder-7B-Instruct}
    \item \texttt{Qwen2.5-Coder-32B-Instruct}
    \item \texttt{microsoft-phi-4}
    \item \texttt{Ministral-3-14B-Instruct-2512}
    \item \texttt{gemma-3-27b-it}
    \item \texttt{Llama-3.1-8B-Instruct}
    \item \texttt{Llama-3.1-70B-Instruct}
\end{itemize}

\section{Reproducibility Statement}
\label{reproduce}
We submit \textbf{1)} the codebase with documentation and \textbf{2)} the full benchmark instances for \textsc{ORCA-Main} and \textsc{ORCA-Project}. Prompt templates are provided in Appendix~\ref{sec:prompt}.

For all experiments, we set the temperature to 0.0 and \texttt{top\_p} to 1.0. Although greedy decoding is theoretically deterministic, run-to-run variance still arises from non-deterministic kernels, batching, and provider-side routing, particularly for proprietary APIs and OpenRouter-served models~\citep{bjarnason2026randomness, atil2024non, ouyang2025empirical, yuan2025understanding}. We therefore run each task three times per model and report the mean Success Rate together with the standard deviation across runs. Proprietary models are evaluated through their official SDKs and APIs, including the Anthropic Python SDK \texttt{anthropic}\footnote{\url{https://platform.claude.com/docs/en/api/client-sdks}} (version 0.11.0) and the Google GenAI SDK \texttt{google-genai}\footnote{\url{https://ai.google.dev/gemini-api/docs/libraries}}. Open-source models with at most 70B parameters are served using \texttt{vLLM}\footnote{\url{https://docs.vllm.ai/en/v0.11.0/}} (version 0.11.0) on 4 NVIDIA H200 GPUs. For larger open-source models, we use the OpenRouter platform\footnote{\url{https://openrouter.ai/}} for inference. Across the full evaluation, local open-source runs consume approximately 200 GPU hours, and total API usage costs approximately USD 1,000.

\section{Limitations}
\label{limitation}
The \textsc{ORCA} benchmark covers 1,600 grounding-level tasks and 200 project-level tasks. Running comprehensive evaluations across multiple models therefore introduces a modest carbon footprint. While we made efforts to optimize our experimental setup, evaluating large proprietary models like \texttt{Claude-Opus-4.6} and \texttt{Gemini-3.1-Pro} requires significant energy consumption that could be reduced with more efficient evaluation protocols in future work.

\section{Ethics Statement}
\label{ethics}
In this work, all benchmark data were derived from publicly available sources, ensuring transparency and reproducibility. We took care to avoid using sensitive or personally identifiable information, and the resulting output remains in the form of programming code, rather than natural language text that might carry an increased risk of biased or harmful content. Our methodology emphasized data integrity, and we reviewed the benchmark data to ensure that they did not contain politically sensitive or ethically concerning material. Throughout the project, we engaged only in model inference without additional training, thereby minimizing both the risk of unintended bias amplification
and the environmental impact associated with extensive computational resources.

\section{Broader Impact}
\label{broader_impact}

\textsc{ORCA} offers an open benchmark that can help the community pinpoint failure modes in data science code translation, guide the development of safer tooling, and reduce the manual effort required when analysts migrate workflows across libraries and back ends. Since our tasks are exclusively in the programming domain and involve only code translation between established libraries, we believe the risk of negative social impact is relatively low.

\onecolumn

\section{Prompt Templates} \label{sec:prompt}

\subsection{Grounding-Level Prompts}

\subsubsection*{Baseline Prompt}
\begin{figure}[H]
    \centering
    \includegraphics[scale=0.9]{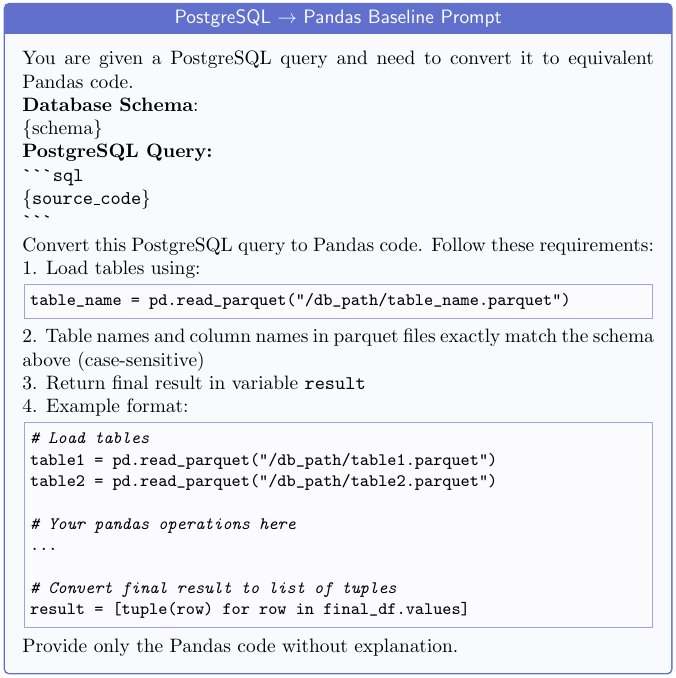}
    \label{fig:prompt_ground_baseline}
\end{figure}
\newpage

\subsubsection*{Intent Prompt (Step 1: Intent Acquisition)}
\begin{figure}[H]
    \centering
    \includegraphics[scale=0.9]{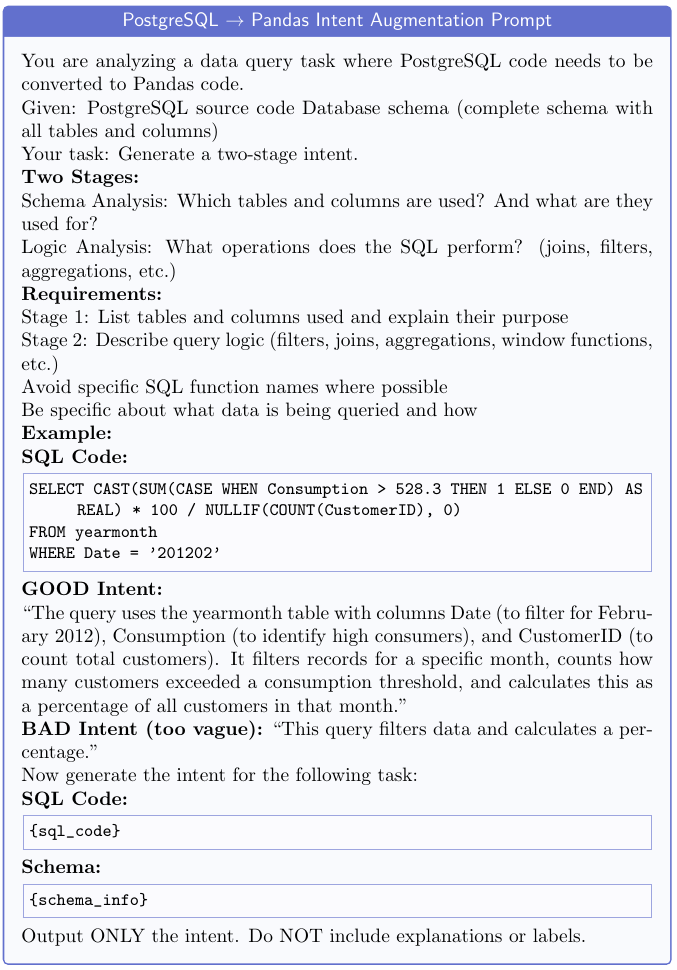}
    \label{fig:prompt_ground_intent_step1}
\end{figure}
\newpage

\subsubsection*{CoT-Intent Prompt (Step 1: Intent Acquisition)}
\begin{figure}[H]
    \centering
    \includegraphics[scale=0.9]{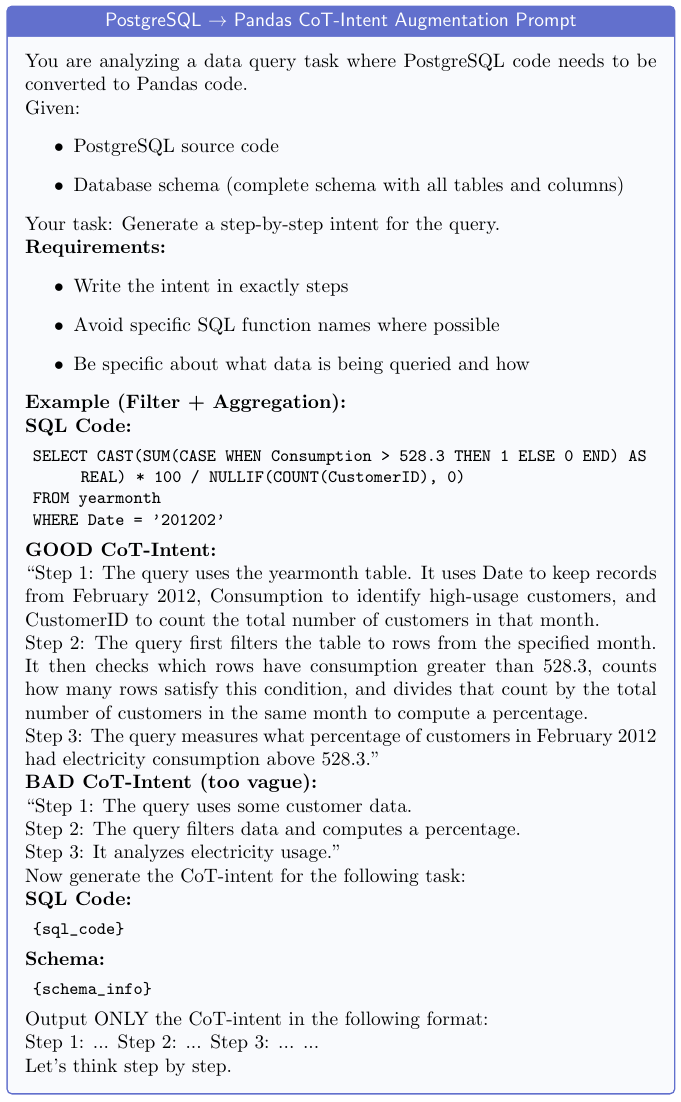}
    \label{fig:prompt_ground_cot_intent_step1}
\end{figure}
\newpage

\subsubsection*{Shared Prompt (Step 2: Translate with Intent)}
\begin{figure}[H]
    \centering
    \includegraphics[scale=0.9]{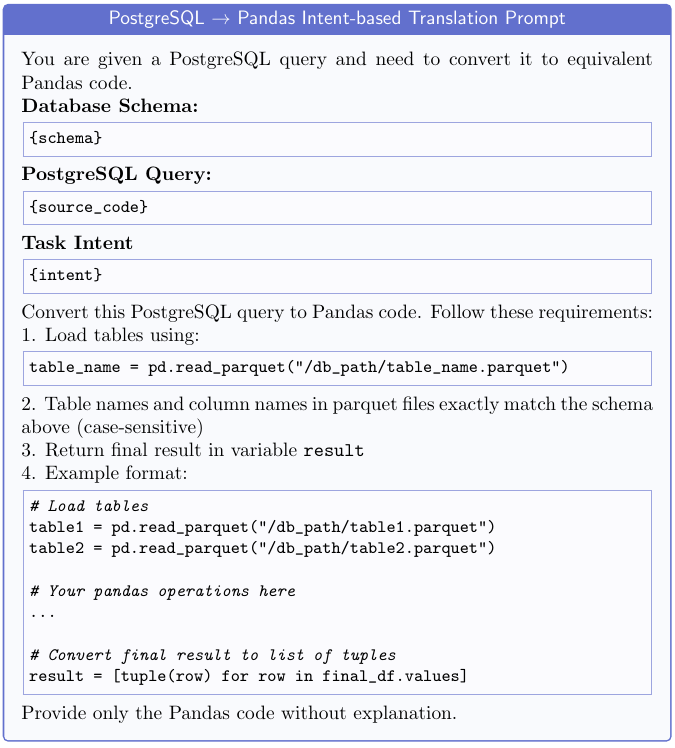}
    \label{fig:prompt_ground_intent_step2}
\end{figure}
\newpage

\subsection{Project-Level Prompts}

\subsubsection*{Action Generation Prompt}
\begin{figure}[H]
    \centering
    \includegraphics[scale=0.9]{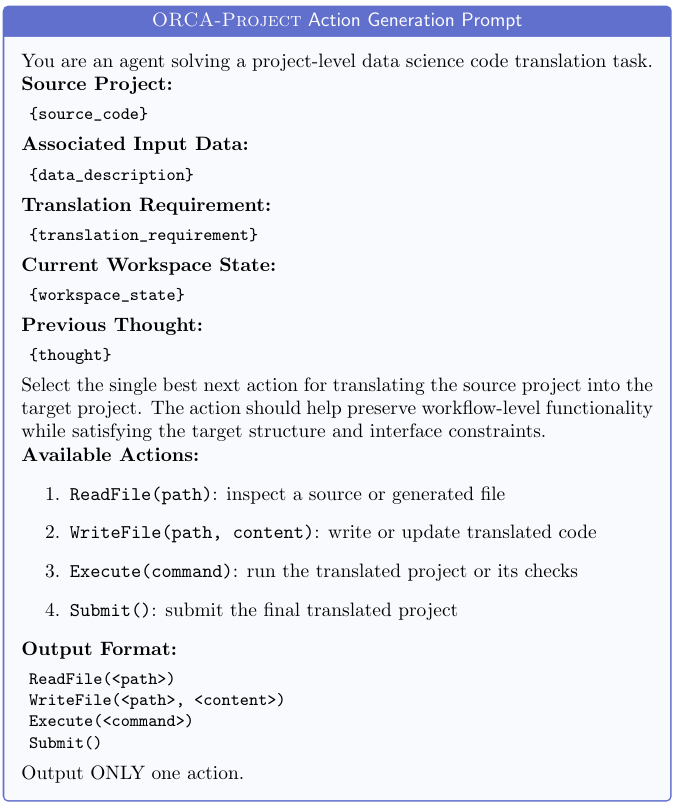}
\end{figure}
\newpage

\subsubsection*{Thought Generation Prompt}
\begin{figure}[H]
    \centering
    \includegraphics[scale=0.9]{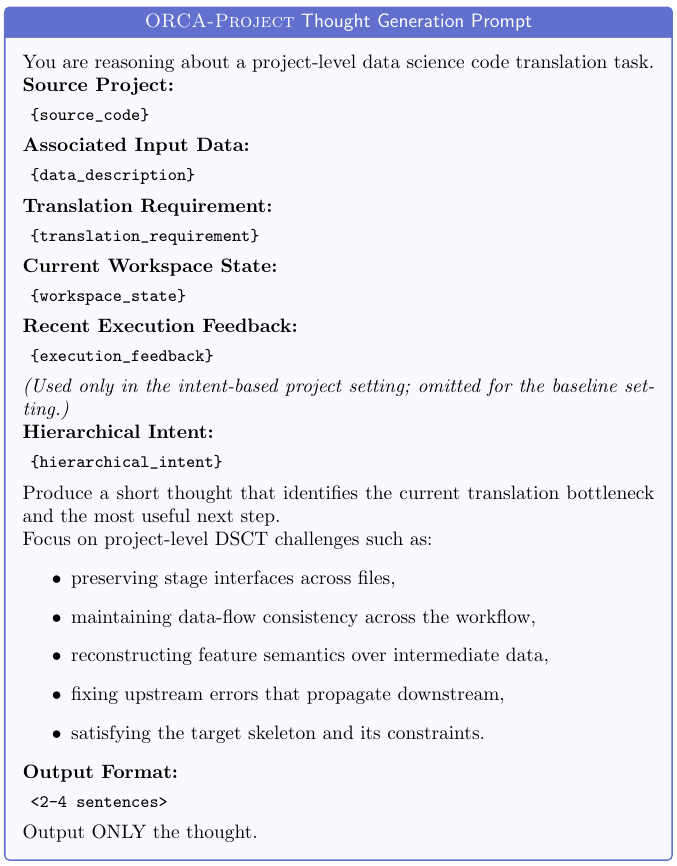}
\end{figure}
\newpage

\subsubsection*{Hierarchical Intent Generation Prompt}
\begin{figure}[H]
    \centering
    \includegraphics[scale=0.9]{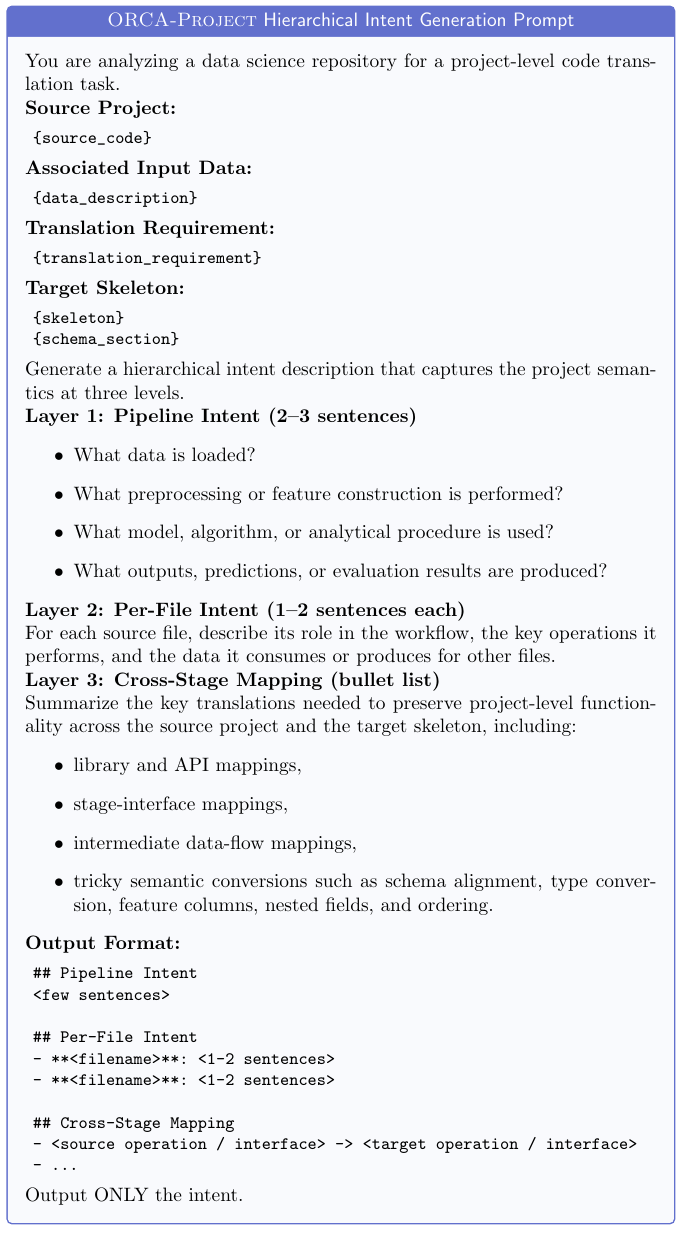}
\end{figure}

\newpage
\subsubsection*{Hierarchical CoT-Intent Generation Prompt}

\begin{figure}[H]
    \centering
    \includegraphics[scale=0.9]{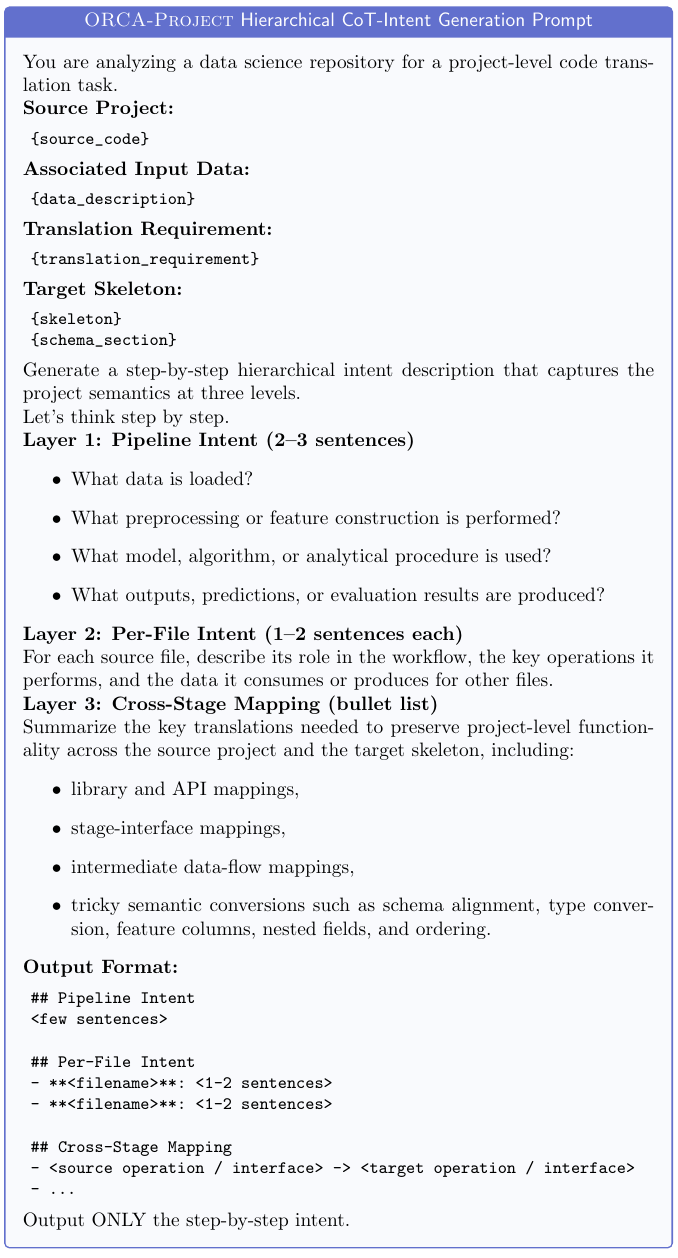}
\end{figure}

\newpage

\end{document}